\documentclass[final,5p,times,twocolumn]{elsarticle}
\usepackage{booktabs}
\usepackage{url}
\usepackage[ruled]{algorithm}  
\usepackage{algpseudocode}  
\usepackage{amsmath}
\usepackage{graphicx} 
\usepackage{subfigure} 
\usepackage{multirow}
\usepackage{multicol}
\usepackage{stfloats}
\usepackage{mathrsfs}
\usepackage{amssymb,amsmath,amsthm}
\usepackage{amsbsy}
\usepackage{stfloats}
\usepackage{nomencl}
\usepackage{tabularx}
\usepackage{bbding}
\usepackage[figuresright]{rotating}
\usepackage{lscape}
\usepackage{pifont}

\usepackage[dvipsnames, svgnames, x11names]{xcolor}
\usepackage{colortbl}  
\usepackage{xcolor}
\usepackage{array}  

\usepackage[caption=false,font=normalsize,labelfont=sf,textfont=sf]{subfig}
\journal{Applied Energy}
\makenomenclature

\usepackage{etoolbox}
\renewcommand\nomgroup[1]{%
	\item[\bfseries
	\ifstrequal{#1}{A}{Acronyms}
	{\ifstrequal{#1}{V}{Variables}{\ifstrequal{#1}{D}{Abbreviations of Methods}}}
	]}
\begin{document}
	\begin{frontmatter}
		\title{ScenGAN: Attention-Intensive Generative Model for Uncertainty-Aware Renewable Scenario Forecasting} 
		%% use optional labels to link authors explicitly to addresses:
		\cortext[correspondingauthor]{Correspondence to: Bo Wang, School of Management and Engineering, Nanjing University, Nanjing 210093, China. E-mail address: bowangsme@nju.edu.cn.}
		\author[label1]{Yifei Wu}
		\author[label2]{Bo Wang*}
		\author[label2]{Jingshi Cui}
		\author[label3]{Pei-chun Lin}
		\author[label4]{Junzo Watada}
		\address[label1]{School of Data Science, Chinese University of Hong Kong - Shen Zhen, Shen Zhen 518172, China}
		\address[label2]{School of Management and Engineering, Nanjing University, Nanjing 210093, China}
		\address[label3]{Department of Information Engineering and Computer Science, Feng Chia University, Taichung City 40710, Taiwan}
		\address[label4]{Graduate School of Information, Production and
			Systems, Waseda University, Kitakyushu 808-0135, Japan}
		
		% The paper headers
		%\markboth{IEEE Transaction on sustainable energy, July~2022}%
		%{Shell \MakeLowercase{\textit{et al.}}: A Sample Article Using IEEEtran.cls for %IEEE Journals}
		
		%\IEEEpubid{0000--0000/00\$00.00~\copyright~2021 IEEE}
		% Remember, if you use this you must call \IEEEpubidadjcol in the second
		% column for its text to clear the IEEEpubid mark.
		
		\begin{abstract}
			To address the intermittency of renewable energy source (RES) generation, scenario forecasting offers a series of stochastic realizations for predictive objects with superior flexibility and direct views. Based on a long time-series perspective, this paper explores uncertainties in the realms of renewable power and deep learning. Then, an uncertainty-aware model is meticulously designed for renewable scenario forecasting, which leverages an attention mechanism and generative adversarial networks (GANs) to precisely capture complex spatial-temporal dynamics. To improve the interpretability of uncertain behavior in RES generation, Bayesian deep learning and adaptive instance normalization (AdaIN) are incorporated to simulate typical patterns and variations. Additionally, the integration of meteorological information, forecasts, and historical trajectories in the processing layer improves the synergistic forecasting capability for multiscale periodic regularities. Numerical experiments and case analyses demonstrate that the proposed approach provides an appropriate interpretation for renewable uncertainty representation, including both aleatoric and epistemic uncertainties, and shows superior performance over state-of-the-art methods.

		\end{abstract}
		
		% This is key words
	\end{frontmatter}
	
	\section{Introduction}
	In response to carbon peaking and carbon neutrality initiatives and sustainable forms of electricity have attracted attention from the energy research community \cite{science}. However, the inherent intermittency of renewable energy source (RES) generation, e.g., the variability and randomness of wind and photovoltaic (PV) power, imposes great uncertainties on power system planning, including medium-term unit commitment \cite{NAVARRO2024122554}, adaptive economic dispatch \cite{wang2024adaptive}, optimal power flow \cite{gao2024bayesian}, and distribution-level market clearing \cite{wang2023tri}. Because of these uncertainties, operators must change the regular modes of conventional grid decision-making and risk assessment. As a dependable technique, uncertainty-aware renewable power forecasting can be used to alleviate the problem of renewable energy curtailment, further exploiting the value of RESs on the generation and distribution sides of power systems.
	
	Nevertheless, mainstream commercial power companies only make static point forecasting results based on physical or statistical calculations publicly available, and these results do not consider the volatile behavior and uncertain characteristics that are often found in real-world operations. To address these issues, power forecasts accommodating uncertainties have been extensively emphasized in relevant studies. Given the different focuses of practical research, uncertainty representation takes various forms in renewable power forecasting, including risk indices \cite{review1}, prediction intervals \cite{d1}, probability densities \cite{ZHANG2020115600}, and spatial-temporal scenarios \cite{s14}. Thereinto, the risk index is meant to give expected information on the forecasting error, prediction intervals cover the upper- and lower-bound levels of future possible values with a certain confidence degree, and density forecasts provide the probability density distribution for the profiles of given periods. However, these three forms all have inherent defects, including the unintuitive risk index, the excessive conservatism of the prediction interval, and the high complexity of the density distribution \cite{survey1}, which limit their effectiveness in uncertainty-aware systems.
	
	As a discrete alternative to the stochastic process of RES generation, scenarios are represented as a set of time trajectories with correlations, which describe the spatial-temporal structure with a long lead time and provide flexible uncertainty qualifications for stochastic programming problems \cite{2020review}. Currently, the conventional methods of scenario generation are mainly divided into three parts: 
	sampling methods based on distribution construction \cite{construction}, reduction methods based on optimization \cite{cluster}, and data-driven methods \cite{s1}. As a specific mode of data-driven scenario generation, \textit{scenario forecasting} to the methods of conducting probabilistic forecasts for a designated period (e.g., day-ahead forecasting) by generating a group of stochastic but predictive scenarios \cite{s2} in a nonparametric
	\begin{table*}[t]
		\caption{Comparison of Eighteen Existing DGMs-based Scenario Generation Studies}
		\renewcommand{\arraystretch}{1.0}
		\begin{center}
			
			\setlength{\tabcolsep}{1.2mm}{
				\begin{tabular}{ccccccccccccccccccc}
					\toprule[0.5mm]
					Ref. & \cite{s1} & \cite{s2} & \cite{s3} &\cite{s4} & \cite{s5} & \cite{s6} & \cite{s7} & \cite{s8}& \cite{s9}& \cite{s10}& \cite{s11} & \cite{s12} & \cite{s13}&\cite{s17}&\cite{s18}& \cite{s15} & \cite{s16} & Ours\\
					\midrule
					
					Information Type & \ding{172}\ding{173} & \ding{172}\ding{173}  & \ding{172} & \ding{172}\ding{173}  & \ding{174} & \ding{172}\ding{173}  & \ding{172}  & \ding{172}\ding{173}  & \ding{172}\ding{173}  & \ding{172}\ding{173} & \ding{172}\ding{173} & \ding{172} & \ding{172}& \ding{172}\ding{173}  & \ding{172}\ding{173}\ding{176}  & \ding{173}\ding{175}  & \ding{172}\ding{173}\ding{175}  & \ding{172}\ding{173}\ding{175} \\

					\midrule
					Uncertainty-aware & \Checkmark &\Checkmark &\Checkmark &\Checkmark &\Checkmark &\Checkmark &\Checkmark &\Checkmark &\Checkmark &\Checkmark &\Checkmark &\Checkmark &\Checkmark &\Checkmark &\Checkmark &\Checkmark &\Checkmark &\Checkmark \\
					Hi-Fi & \XSolidBrush & \XSolidBrush & \XSolidBrush & \XSolidBrush & \XSolidBrush & \XSolidBrush & \XSolidBrush & \XSolidBrush& \XSolidBrush & \XSolidBrush & \XSolidBrush &  \XSolidBrush & \Checkmark&\Checkmark & \Checkmark & \Checkmark & \Checkmark & \Checkmark\\
					Forecastable & \XSolidBrush & \Checkmark & \Checkmark & \XSolidBrush & \XSolidBrush & \XSolidBrush & \Checkmark & \Checkmark& \XSolidBrush & \XSolidBrush & \XSolidBrush &  \Checkmark & \Checkmark& \Checkmark & \Checkmark & \Checkmark & \Checkmark & \Checkmark\\
					Meteorologic & \XSolidBrush & \XSolidBrush& \XSolidBrush & \XSolidBrush &  \Checkmark & \XSolidBrush &  \XSolidBrush &  \XSolidBrush & \XSolidBrush & \XSolidBrush & \XSolidBrush &   \XSolidBrush & \XSolidBrush & \XSolidBrush & \XSolidBrush & \Checkmark & \Checkmark & \Checkmark\\
					End-to-End & \XSolidBrush & \XSolidBrush  & \XSolidBrush  & \XSolidBrush & \XSolidBrush & \XSolidBrush &\XSolidBrush & \XSolidBrush & \XSolidBrush & \XSolidBrush &\XSolidBrush & \Checkmark &\XSolidBrush & \XSolidBrush & \XSolidBrush  & \XSolidBrush & \XSolidBrush  & \Checkmark\\
					Epistemic-aware & \XSolidBrush & \XSolidBrush  & \XSolidBrush  & \XSolidBrush & \XSolidBrush & \XSolidBrush &\XSolidBrush & \XSolidBrush & \XSolidBrush & \XSolidBrush &\XSolidBrush & \XSolidBrush &\XSolidBrush & \XSolidBrush & \Checkmark  & \XSolidBrush & \XSolidBrush  & \Checkmark\\
					\bottomrule[0.5mm]	
					
			\end{tabular}}
			\label{tab1}
		\end{center}
		\Checkmark: demand fulfillment. \XSolidBrush: demand unfulfillment. \textbf{Information Type}: information types involved in the methods. \textbf{Uncertainty-aware}: uncertainty is considered. \textbf{Hi-Fi}: high-fidelity. \textbf{Forecastable}: generated scenarios possess predictive characteristic. \textbf{Meteorologic}: meteorological factors are taken into account. 
		\textbf{End-to-End}: scenarios with predictive ability are generated by one step.
		\textbf{Epistemic-aware}: epistemic uncertainty is considered. \textbf{\ding{172}}: wind power. \textbf{\ding{173}}: PV power. \textbf{\ding{174}}: solar irradiance. \textbf{\ding{175}}: numerical weather prediction (NWP). \textbf{\ding{176}}: load.
		
	\end{table*}  
	 and distribution-free manner; these methods impose higher requirements on variable dimensionality, forecast accuracy and pattern adequacy. To address these challenges, current scenario forecasting methods mainly focus on the machine learning (ML) field, especially by using neural network techniques \cite{YE20231091} to fit the complex shapes of future renewable generation.  
	
	Due to the advantages of adaptive generation, nonsupervision and fast expansion, the newest scenario generation studies (mostly scenario forecasting studies) have focused on deep generative models (DGMs), especially generative adversarial networks (GANs). Along this line, a review of the existing DGM-based scenario generation research is presented in Table 1, which also shows a forward-looking comparison with this paper's work to highlight its superiority. Below, we highlight some of the issues worth investigating.
	
	First, although existing studies have demonstrated the fair performance of DGMs in scenario generation, as Table 1 reveals, almost none of the DGM-based methods present a systematic analysis of the determination of renewable uncertainties, or haven't been focus on the mechanisms and guiding methods behind the uncertainties. Second, the existing modeling paradigm still has great room for improvement in the field of scenario forecasting. Most of these methods follow a two-stage procedure, i.e., a generative model is proposed first to construct mappings from latent variables to sample distributions, while the fixed model is further guided to generate forecasts by adopting optimization \cite{s17}, cluster \cite{s18}, or auxiliary networks \cite{s16}. These complicated procedures make the generated scenario's effectiveness overly dependent on the capability of first-stage models, often leading to inefficient capture of future dynamic behaviors or deficiency in covering forecasting distributions. In addition, cascading failure is a nonnegligible problem. Finally, transformer (an attention-based network) \cite{transformer} and its variants \cite{informer} have achieved great success in sequence-to-sequence generative tasks, which also opens up new possibilities for capturing long time-series patterns and spatiotemporal correlations in scenario forecasting.
	
	% including deep convolutional GAN (DCGAN) \cite{s1}, DCGAN followed by stochastic optimization (DCGAN-SO) \cite{s2}, multi agent diverse GAN with stochastic optimization (MADGAN-SO) \cite{s3}, Bayesian GAN \cite{s4}, conditional GAN (CGAN) \cite{s5}, improved GAN combined with variational inference (GAN-VI) \cite{s6}, conditional Wasserstein GAN with gradient penalty (C-WGAN-GP) \cite{s7}, Wasserstein GAN with gradient penalty, consistency term and stochastic optimization (WGAN-GP-CT-SO) \cite{s8}, InfoGAN \cite{s9}, controllable GAN (Ctrl-GAN) \cite{s10}, least square GAN with federated learning (Fed-LSGAN) \cite{s11}, sequence GAN based on long short-term memory (SeqGAN-LSTM) \cite{s12}, progressive growing of GAN followed by multiobjective optimization (ProGAN-MO) \cite{s13}, a Gramian angular field-based GAN model called StyleGAN-ADA-ESR-SO \cite{s17}, long short-term memory-based auto-encoder (LSTM-AE) \cite{s15}, networks combined with bidirectional long-short term memory, convolutional neural network, and gated recurrent unit (BiLSTM-CNN-GRU) \cite{s14}, and conditional style-based GAN followed by sequence encoder (C-StyleGAN2-SE) \cite{s16}.
	
	In response to the above discussion, a novel scenario forecasting model named Scenario GAN (ScenGAN) is proposed based on GANs and Bayesian theory to explore the feasible forecasting distribution of renewable power, as shown in Fig. 1. The major contribution promoted by ScenGAN at the cutting edge of DGM-based scenario generation, which differs with previous works presented in Table 1, is summarized in three parts:
	\begin{figure*}[h]
		\centerline{\includegraphics[width=18cm]{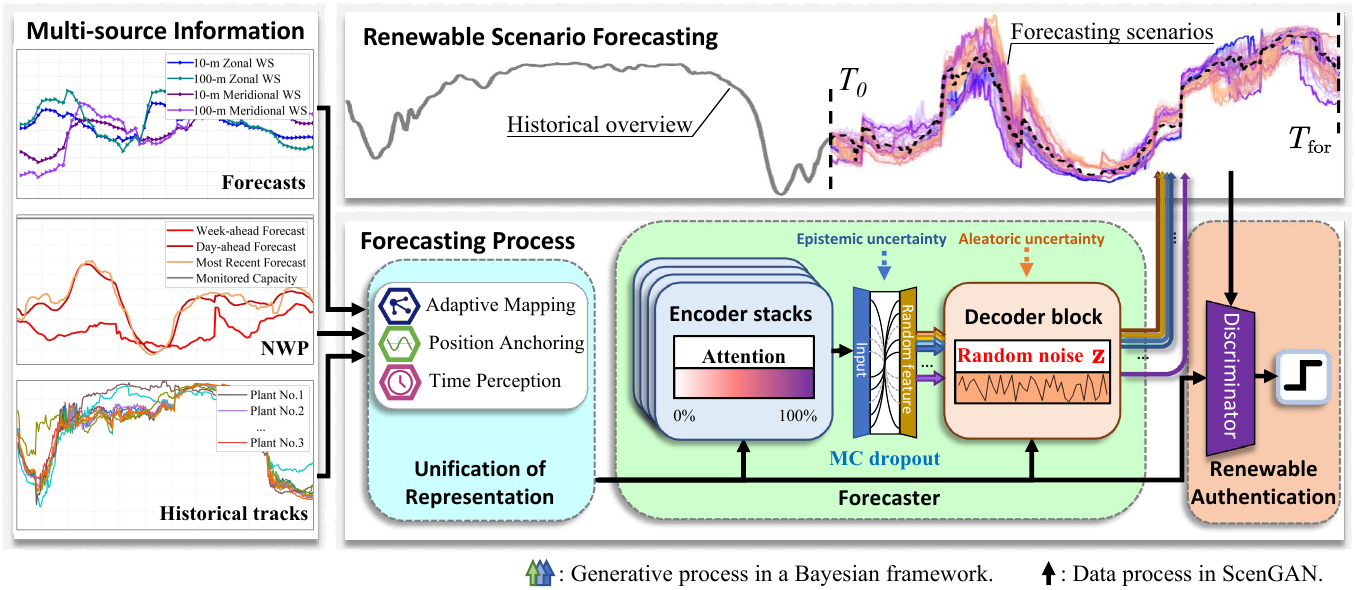}}
		\caption{ Overall framework for the task of renewable scenario forecasting.}
		\label{fig}
	\end{figure*}
	\subsection{Handling Uncertainties in Scenario Forecasting} 
	Under the deep Bayesian architecture, a scenario-based probabilistic forecasting method is proposed to make
	full use of multiple source information, and overcome the capacity limitations of conventional scenario generation networks, giving the predictive scenarios the capability of accounting for both epistemic and aleatoric uncertainty. 
	
	\subsection{Attention-Intensive Generative Modeling} 
	Given the superiority of transformer in capturing long-range interactions and generative models in generating realistic artifacts, a modified deep generative model incorporating an attention-based mechanism and adversarial architecture is proposed to capture complex spatial-temporal dependencies, renewable authenticity (multiscale regularities
	 of renewable power curves), and the intrinsic coupling characteristics of multiple meteorological types, outperforming the state-of-the-art methods in scenario forecasting.
	
	\subsection{An End-to-End Scenario Forecasting Model}  
	Although ScenGAN performs complex forecasting, it has a minimalist end-to-end prediction paradigm, which enables it to perform well on multiple datasets with similar granularities. Moreover, ScenGAN has also been shown to have considerable adaptability in day-ahead power system operations, proving that its simple operation and strong generation capabilities can provide new solutions for industrial applications.
	
	The remainder of this paper is organized as follows: Section II introduces the preliminary information. Section III manifests the ScenGAN scenario forecasting model. Section IV presents the experimental results. Finally, Section V draws conclusions.
	\section{Preliminary}
	\subsection{Renewable Uncertainties}
	Modelers are primarily interested in the flow of uncertainty from inputting variables to forecasting distributions, which is also critical in the case of renewable power forecasting that incorporates inherent intermittency, i.e., renewable uncertainties. First, renewable energy is inherently uncertain. Taking wind energy as an example, Fig. 2 depicts a simplified diagram of the wind spectrum, where local peaks indicate significant changes and fluctuations in wind speed over the corresponding time period. The figure intuitively shows that uncertainties occur to varying degrees at multiple time scales in the actual power production process, e.g., short-term responses to atmospheric turbulence effects or long-term climate changes caused by human activities or macrometeorological factors. Second, the limitations of forecast methods can also lead to suboptimal and uncertain results \cite{distribution}. From the perspective of data-driven modeling, the intermittent nature of the RES generation process can be further interpreted in terms of \textit{aleatoric uncertainty} and \textit{epistemic uncertainty} \cite{d2}. 
	\begin{figure}[h]
		\centerline{\includegraphics[width=9cm]{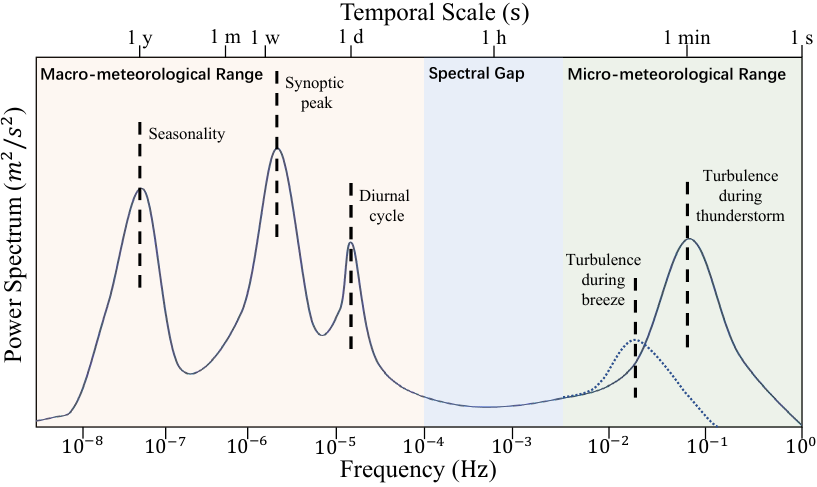}}
		\caption{ Exemplification of wind spectrum depicts typical behavior patterns.}
		\label{fig}
	\end{figure}
	\subsubsection{Aleatoric Uncertainty}
	In this field, aleatoric uncertainty is conventionally used to describe the intrinsic noise of renewable power data due to multiple affective factors, such as climate change, turbulence, and aperiodic human activity. Developing a DGM-based scenario generation model can be seen as an effective solution, as it addresses the massive aleatoric uncertainty in the RES generation process and provides a set of reliable scenarios to fit the possible forecasting distributions.
	\subsubsection{Epistemic Uncertainty}
	The term epistemic uncertainty refers to the uncertainty in the model architecture due to modeling limitations. In addition to aleatoric uncertainty, epistemic uncertainty is a pivotal part of renewable scenario generation that reveals the overall uncertainty degree of the scenarios inferred from a certain forecasting model. For DGMs, epistemic uncertainty is widely found across possible network structures and parameter settings. It is important to understand the extent to which the selected model can accurately forecast power behavior with different information (e.g., point forecasts, historical tracks, and weather information), as capturing epistemic uncertainty can theoretically result in more robust decision-making in situations of insufficient information or complex patterns.
	\subsection{Problem Formulation}
	The main objective is to reveal inherent renewable uncertainties driven by the aforementioned factors and effectively incorporate them into renewable scenario forecasting. Denoting the available informational scalar and historical power values as $\mathring{x}_t$ and $\mathrm{x}_t$ at time node $t$, we integrate $N_T$-consecutive-step timelines before node $T$ as the informational profile $\mathring{\boldsymbol{x}}_{T:N_T}$ and the actual profile $\mathbf{x}_{T:N_T}$:
	\begin{equation}
		\begin{aligned}
			\mathring{\boldsymbol{x}}_{T:N_T} = [\mathring{x}_{T-N_T+1}, ..., \mathring{x}_{T-1}, \mathring{x}_{T}], 
		\end{aligned}
	\end{equation}
	\begin{equation}
		\begin{aligned}
			\mathbf{x}_{T:N_T} = [\mathrm{x}_{T-N_T+1}, ..., \mathrm{x}_{T-1}, \mathrm{x}_{T}].
		\end{aligned}
	\end{equation}
	
	Accounting for the $N_\rho$-site relationship,  profiles can be further integrated as two forms of multivariable samples, i.e., informational samples  $X_{T:N_T}^{\mathrm{info}}$ and historical samples $X_{T:N_T}^{\mathrm{hist}}$: 
	\begin{equation}
		\begin{aligned}
			X_{T:N_T}^{\mathrm{info}}=\{[\mathring{\boldsymbol{x}}^{(1)}_{T:N_T}, \mathring{\boldsymbol{x}}^{(2)}_{T:N_T}, ..., \mathring{\boldsymbol{x}}^{(N_\rho)}_{T:N_T}]^\intercal| \mathring{\boldsymbol{x}}^{(i)}_{T:N_T} \in \mathcal{X}_{\mathrm{info}}\},
		\end{aligned}
	\end{equation}
	\begin{equation}
		\begin{aligned}
			X_{T:N_T}^{\mathrm{hist}}=\{[\mathbf{x}^{(1)}_{T:N_T}, \mathbf{x}^{(2)}_{T:N_T}, ..., \mathbf{x}^{(N_\rho)}_{T:N_T}]^\intercal| \mathbf{x}^{(i)}_{T:N_T} \in \mathcal{X}_{\mathrm{hist}}\},\,
		\end{aligned}
	\end{equation}
	where $\mathcal{X}_{\mathrm{info}}$ and $\mathcal{X}_{\mathrm{hist}}$ indicate the informational and historical datasets, respectively. 
	
	In practical modeling, Bayesian deep learning extends vanilla deterministic networks to a posterior inference, where the model parameters and the corresponding outputs are assumed to be distributions rather than deterministic point values, resulting in an inference profile that captures both aleatoric and epistemic uncertainties. In this way, the interpretable part of the uncertainties is controlled by prior distributions on the network parameter $\theta$, while the aleatoric part is measured by the noise of the forecasted outputs $\hat{X}_{T:N_T}$ of a deterministic generative model. Given the input $X_{T:N_T}=\{X_{T:N_T}^{\mathrm{info}}, X_{T:N_T}^{\mathrm{hist}}\}$, the probability distribution of $\hat{X}_{T_{\mathrm{for}}:N_T}$ with the lead time node $T_\mathrm{for} (T_\mathrm{for}>T)$ is:
	\begin{equation}
		\begin{aligned}
			P\!(\hat{X}\!_{T_{\mathrm{\!for}}\!:N\!_T}|X\!_{T:N\!_T}, \mathbf{D})=\int \!\!P\!(\hat{X}\!_{T_{\mathrm{\!for}}\!:N\!_T}|X\!_{T:N\!_T}, \theta)P\!(\theta| \mathbf{D})d\theta, \\
		\end{aligned}
	\end{equation}
	where $P\!(\theta| \mathbf{D})$ is the posterior distribution, which represents the epistemic uncertainty over a given dataset $\mathbf{D}=\{\mathcal{X}_{\mathrm{hist}}, \mathcal{X}_{\mathrm{info}}\}$.
	
	Ultimately, the task of renewable scenario forecasting for the time node $T_\mathrm{for}$ can be formulated as a maximum likelihood estimation:
	\begin{equation}
		\begin{aligned}
			\{\hat{X}_{T_{\mathrm{\!for}}\!:N\!_T}^*\}_{N\!\!_fN\!\!_n}\!=\!\mathop{\arg\max}_{\{\hat{X}_{T_{\mathrm{\!for}}\!:N\!_T}\}}\prod_{i=1}^{N_f}\prod_{j=1}^{N_n}\!\!P_{i, j}(Y_{T_{\mathrm{\!for}}\!:N\!_T}|X\!_{T:N\!_T}, \mathbf{D}),
		\end{aligned}
	\end{equation}
	where $\{\hat{X}_{T_{\mathrm{\!for}}\!:N\!_T}^*\}_{N_fN_n}$ denotes $N_f\times N_n$ optimal scenarios that describe $N_f$ possible patterns representing epistemic uncertainty, where each pattern has corresponding uninterpretable noise; $Y_{T_{\mathrm{\!for}}\!:N\!_T}$ represents the actual observations.
	\section{ScenGAN Scenario Forecasting Model}
	In this section, ScenGAN is proposed for renewable scenario forecasting, as shown in Fig. 3. First, a data preprocessing method is described. Second, an encoder-decoder subnetwork named \textit{Forecaster} is designed to produce accurate forecasts. Finally, the framework, uncertainty processing and loss function of ScenGAN are introduced in detail.
	
	\subsection{Renewable Information  PreProcessing}
	For full access to information, the inputs of ScenGAN consist of three components: the encoder input $X^{\mathrm{enc}}_{T:N_T}$, decoder input $X^{\mathrm{dec}}_{T_\mathrm{for}:N_T}$, and discriminator input $X^{\mathrm{dis}}_{\hat{T}:N_T}$, as shown in Fig. 3(a). Their processing methods are described in this subsection.
	\subsubsection{Multilevel Embedding for Forecaster Input}In scenario forecasting, the local behavior, diurnal patterns and long-term seasonality of the RES generation indicate multilevel inherent correlations \cite{s16}. This motivates us to design time-stamp embedding for RES generation trajectories.
	
	First, a fixed sine-cosine positional embedding $\mathrm{PE}(\cdot)$ is used to extract local orders:
	\begin{equation}
		\mathrm{PE}(t, i) =\left\{
		\begin{aligned}
			&\!\sin(t/10000^{i/N_m}), i\!=\!2k, k\!\in\!\mathbb{N}\\
			&\!\cos(t/10000^{(i-1)/N_m}), i\!=\!2k\!+\!1, k\!\in\!\mathbb{N}\\
		\end{aligned}
		\right.,
	\end{equation}
	where $N_m$ indicates the feature dimension after processing and $i\!\!\in\!\!\{\!1, 2, ...\lfloor N_m/2 \rfloor\!\}$. 
	As mentioned in Subsection II-B, the historical sample is denoted as $X_{T:N_T}^{\mathrm{hist}}$; the informational sample $X_{T:N_T}^{\mathrm{info}}$ is obtained from the weather records and available point forecasts or historical tracks; and the global timestamp includes the \textit{year} $x^\mathrm{y}$, \textit{month} $x^\mathrm{m}$, \textit{day} $x^\mathrm{d}$,  \textit{hour} $x^\mathrm{h}$, \textit{minute} $x^\mathrm{min}$, and \textit{typical event} $x^\mathrm{eve}$. A stamp embedding $\mathrm{SE}(\cdot)$ is employed to extract the global context:
	\begin{equation}
		\begin{aligned}
			\setlength{\arraycolsep}{1.0pt}
			\mathrm{SE}(t)=\mathrm{Embedding}(x^\mathrm{y}_t, x^\mathrm{m}, x^\mathrm{d}, x^\mathrm{h}, x^\mathrm{min}, x^\mathrm{eve}),
		\end{aligned}
	\end{equation}
	where $\mathrm{Embedding}(\cdot)$ is the symbolic representation of  the embedding layer. 
	
	In addition, the learnable projection operations are added to align the feature dimensions, and the inputs $X^{\mathrm{enc}}_{T:N_T}$ and $X^{\mathrm{dec}}_{T_{\mathrm{\!for}}:N_T}$  are integrated for Forecaster as follows:
	\begin{equation}
		\begin{aligned}
			\,\,\,\,	X^{\mathrm{enc}}_{T:N_T}\!=\!\big[\!\big[ \mathrm{u}_t^{(i)}\!+\!\mathrm{PE}(t, i)\!+\!\mathrm{SE}(t)\big]^{T}_{t=T\!-\!N_T\!+\!1}\!\big]^{\!N_m}_{i=1},\,\,\,
		\end{aligned}
	\end{equation}
	\begin{equation}
		\begin{aligned}
			\quad\,\,\,\,	X^{\mathrm{dec}}_{T_{\mathrm{\!for}}:N_T}\!=\!\big[\!\big[ \mathrm{v}_t^{(i)}\!+\!\mathrm{PE}(t, i)\!+\!\mathrm{SE}(t)\big]^{\!T_{\mathrm{\!for}}}_{\!t=T_{\mathrm{\!for}}\!-\!N_T\!+\!1}\!\big]^{\!N_m}_{i=1},
		\end{aligned}
	\end{equation} 
	where $\mathrm{u}$ and $\mathrm{v}$ are scalars obtained by the projection operation:
	\begin{equation}
		\begin{aligned}
			\quad	[[\mathrm{u}]_{N_T}]_{N_m}=\mathrm{LReLU}( X_{T_\mathrm{for}:N_T}^{\mathrm{info}}\!\!\mathrm{W}_1+\mathrm{B}_1),\,
		\end{aligned}
	\end{equation} 
	\begin{equation}
		\begin{aligned}
			\quad	[[\mathrm{v}]_{N_T}]_{N_m}=\mathrm{LReLU}( X_{T_\mathrm{for}:N_T}^{\mathrm{rest}}\!\!\mathrm{W}_2+\mathrm{B}_2),\,
		\end{aligned}
	\end{equation} 
	where, $\small{\mathrm{LReLU}(\cdot)}$ is the leaky rectifier linear unit activation function, $\mathrm{W}_\bullet$ and $\mathrm{B}_\bullet$ are the weights and bias, and $X_{T_\mathrm{for}:N_T}^{\mathrm{rest}}$ is the remaining variable that masks the unknown realizations.
	
	\begin{figure*}[t] 
		\centerline{\includegraphics[width=18cm]{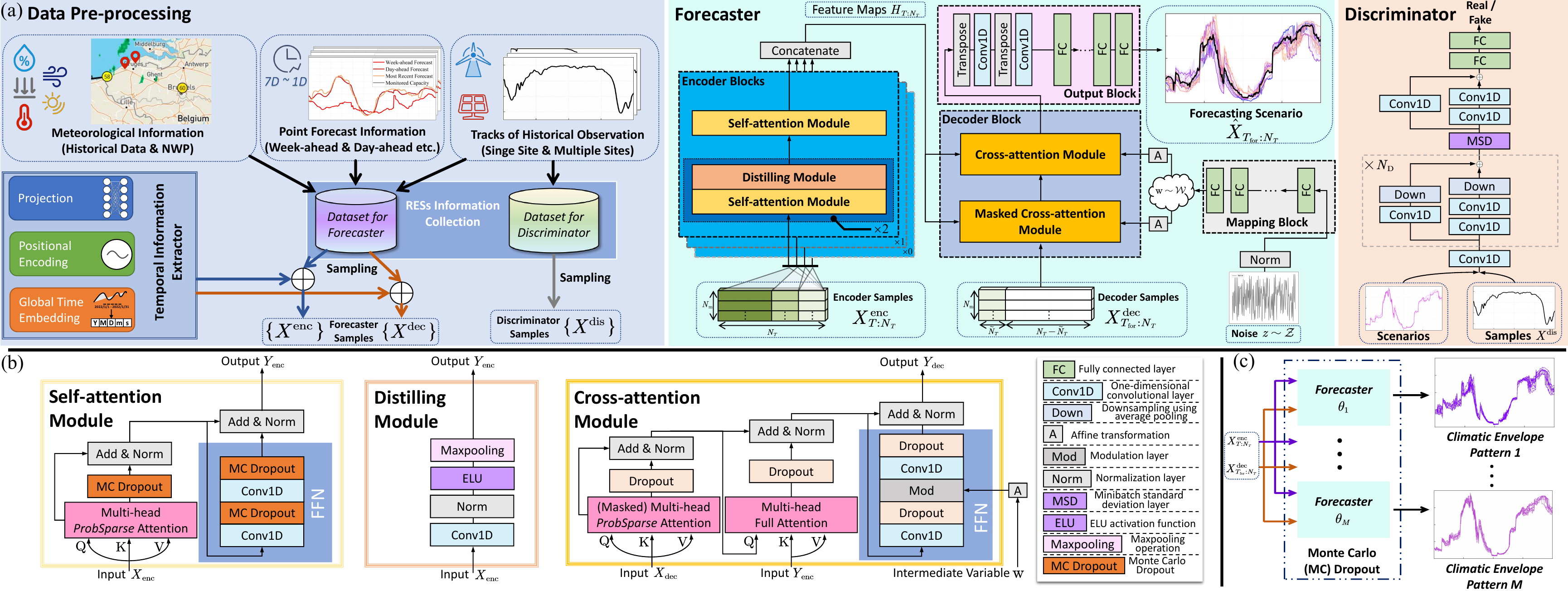}}
		\caption{Perspective drawing of the proposed ScenGAN network for renewable scenario forecasting. Model skeleton (a). Detailed views of three critical modules in attention-intensive Forecaster subnetwork (b). Mechanism of variational dropout for simulating possible day-ahead profiles (c).}
		\label{fig}
	\end{figure*}
	\subsubsection{Representations of the Discriminator Input}
	To control the forecasting authenticity, historical sets $\mathcal{X}_{\mathrm{hist}}$ are applied so that discriminator can mine the inherent patterns: 
	\begin{equation}
		\begin{aligned}
			X_{\hat{T}:N_T}^{\mathrm{dis}}=\{[\mathbf{x}^{(1)}_{\hat{T}:N_T}, \mathbf{x}^{(2)}_{\hat{T}:N_T}, ..., \mathbf{x}^{(N_\rho)}_{\hat{T}:N_T}]^\intercal| \mathbf{x}^{(i)}_{\hat{T}:N_T} \!\!\!\!\in\!\! \mathcal{X}_{\mathrm{hist}}\}.
		\end{aligned}
	\end{equation}
	\subsection{Critical Subnetwork Forecaster for Power Forecasting}
	In the conventional sense, the attention-intensive transformer follows a nonrecursive encoder-decoder structure and positional encoding strategy \cite{transformer}, possessing the ability to capture the temporal dependencies in long time series forecasting (LTSF) tasks. This is mainly attributed to the parallel input forms of long sequences and the persistence of local information in the self-attention mechanism. Essentially, the canonical self-attention mechanism acts as a scaled dot-product:
	\begin{equation}
		\begin{aligned}
			\mathcal{A}(\mathrm{Q}, \mathrm{K}, \mathrm{V})=\mathrm{Softmax}(\mathrm{Q} \mathrm{K}^\intercal/\sqrt{d_k})\mathrm{V},
		\end{aligned}
	\end{equation}
	where, queries $\mathrm{Q}\in\mathbb{R}^{l_Q\times d_k}$, keys $\mathrm{K}\in\mathbb{R}^{l_K\times d_k}$, and values $\mathrm{V}\in\mathbb{R}^{l_K\times d_v}$. $l_Q$ and $l_K$ denote the lengths, $d_k$ and $d_v$ denote the feature dimensions. Furthermore, transformer integrates multihead attention to focus on $h$ aspects of the pattern:
	\begin{equation}
		\begin{aligned}
			\mathrm{MultiHead}=\mathrm{Concat}(\mathrm{head}_1, ...,\mathrm{head}_h)\mathrm{W^O},
		\end{aligned}
	\end{equation}
	\begin{equation}
		\begin{aligned}
			\mathrm{head}_i=\mathcal{A}(\mathrm{Q}\mathrm{W^Q}, \mathrm{K}\mathrm{W^K}, \mathrm{V}\mathrm{W^V}),\,\,
		\end{aligned}
	\end{equation}
	where, $\mathrm{Concat}(\cdot)$ indicates the concatenation operation and $\mathrm{W^O}$, $\mathrm{W^Q}$, $\mathrm{W^K}$,  and $\mathrm{W^V}$ are learnable weight matrices.
	
	Nevertheless, problems such as inefficient inference and complex  architecture limit the forecasting effect of the vanilla transformer. Inspired by the long tail structure of attention, a modified subnetwork Forecaster is designed to handle the LTSF task.
	\subsubsection{ProbSparse Attention Modules}
	Empirically, the self-attention feature map in transformer has a long-tail distribution; i.e., a few dot-product pairs dominate in constructing attention, while others contribute little to the forecasting results. Therefore, it is ideal to focus only on the crucial parts and discard inefficient results. A query's sparsity measurement $\mathrm{SM}(\mathbf{q}_i, \mathrm{K})$ is formulated to extract noteworthy components:
	\begin{equation}
		\begin{aligned}
			\mathrm{SM}(\mathbf{q}_i, \mathrm{K})=\max_j\!\Big\{\mathbf{q}_i\mathbf{k}_j^\intercal/\sqrt{d_k}\Big\}\!-\!(1/l_K)\!\sum_{j=1}^{l_K}\!\!\Big(\mathbf{q}_i\mathbf{k}_j^\intercal/\sqrt{d_k}\Big),
		\end{aligned}
	\end{equation}
	where $\mathbf{k}_j$ is the $j$-th row of $\mathrm{K}$.  In practice, $\mathrm{\hat{K}}$ with $U$ rows is randomly selected to compute $\mathrm{SM}(\mathbf{q}_i, \mathrm{\hat{K}})$. Then, this measurement is sorted to extract the Top-$u$ as sparse queries $\hat{Q}$, and the self-attention operation is reformulated as follows:
	\begin{equation}
		\begin{aligned}
			\mathcal{\hat{A}}(\mathrm{Q}, \mathrm{K}, \mathrm{V})=\mathrm{Softmax}(\mathrm{\hat{Q}} \mathrm{K}^\intercal/\sqrt{d_k})\mathrm{V}.
		\end{aligned}
	\end{equation}
	
	Compared with full self-attention, ProbSparse self-attention improves the computing efficiency without losing any crucial information, and the query abandonment of low scores avoids giving excessive attention to irregular mutations, guiding the model to focus on the overall trend and decisive changes. Moreover, the random sampling strategy leads to query diversity, ensuring that each head can focus on different renewable characteristics, e.g., various forms of ramp events.
	\subsubsection{Other Improvements}  
	To improve the attention efficiency of spatial-temporal features in multivariate conditions, some other techniques are employed to support ProbSparse attention, including \textit{attention distilling} in the encoder, \textit{generative inference} in the decoder, and a \textit{spatial-aware convolutional module} in the output block, as shown in Fig. 3(a). 
	\subsection{ScenGAN for End-to-End Scenario Forecasting}
	Similar to the vanilla GAN framework, ScenGAN consists of two competing components: Forecaster $\mathcal{F}$ (an alternative to the generator) and a discriminator $\mathcal{D}$. According to the adversarial idea \cite{goodfellow}, $\mathcal{F}\!:\!\!\mathcal{Z}\to\mathcal{X}$ tries to produce forecasts that approximate the actual observations, while $\mathcal{D}:\mathcal{X}\to[0,1]$ tries to discriminate the actual and generated samples. On the other hand, what makes $\mathcal{F}$ different from a normal generator is that $\mathcal{F}$ is anchored to a specified forecasting distribution rather than the whole historical overview, and the latent variable $z\sim \mathcal{Z}$ no longer plays a leading role in mapping output patterns but only represents detailed random noise. This redesign is motivated by the fact that, the characteristics of end-to-end scenario forecasting require ScenGAN to be directly forecastable, which means that the generated scenarios should converge to the vicinity of the ground-truth and maintain diversity simultaneously in a single feedforward output.
	\subsubsection{Capturing Uncertainties}
	
	To address these issues, Monte Carlo (MC) dropout and adaptive instance normalization (AdaIN) are adopted to capture renewable uncertainties. 
	
	[\textit{Variational Inference}]: According to Formula (5), the posterior distribution $P(\theta|\mathbf{D})$ that represents epistemic uncertainty is actually intractable in complex networks such as ScenGAN. As an alternative, an explicit variational distribution $P(\theta|\mathrm{W})$ parameterized by fixed parameters $\mathrm{W}$ guarantees that the optimal distribution $P'(\theta|\mathrm{W})$ can characterize $P(\theta|\mathbf{D})$ well by minimizing the Kullback-Leibler (KL) divergence:
	\begin{equation}
		\begin{aligned}
			\mathcal{KL}(P(\theta|\mathrm{W})\|P(\theta|\mathbf{D}))=\int P(\theta|\mathrm{W})\log\frac{P(\theta|\mathrm{W})}{P(\theta|\mathbf{D})}d\theta.
		\end{aligned}
	\end{equation} 
	
	To avoid directly calculating the intractable posterior, it is deduced by applying the Bayes formula as follows:
	\begin{equation}
		\vspace*{-0.1\baselineskip}
		\begin{aligned}
			\mathcal{KL}(P(\theta|\mathrm{W})\|P(\theta|\mathbf{D}))=&\log(P(\mathbf{D}))+\mathcal{KL}(P(\theta|\mathrm{W})\|P(\theta))
			\\	&-\int P(\theta|\mathrm{W})\log{P(\mathbf{D}|\theta)}d\theta,
		\end{aligned}
	\end{equation} 
	where $\int P(\theta|\mathrm{W})\log{P(\mathbf{D}|\theta)}d\theta$ represents the expectation of $P(\mathbf{D}|\theta)$ for $\theta\!\!\sim\!\!P(\theta|\mathrm{W})$, which is known as the data-based likelihood term;  $\mathcal{KL}(P(\theta|\mathrm{W})\|P(\theta))$ is known as the complexity term, determined by the prior distribution; and $\log(P(\mathbf{D}))$ is known as the irrelevant evidence term. In short, the objective of variational inference is viewed as a trade-off between the likelihood expectation and the prior–posterior differences \cite{bayesian1}.
	
	[\textit{MC Dropout}]: Essentially, MC dropout is a Bayesian theoretical tool that expresses epistemic uncertainty through the permanent opening of the dropout layer. To confirm prior beliefs, the MC dropout-connection for network-oriented parameters is incorporated into the encoder block. The random dropout in the encoder perturbs the input intelligently in the feature map space, which leads to potential feature changes and is further propagated through the decoder network. A prior distribution $q(\theta_\mathcal{F})$ on a randomly-zeroed matrix $\Omega$ is employed to approximate the posterior distribution $P(\theta|\mathbf{D})$: 
	\begin{equation}
		\vspace*{-0.2\baselineskip}		
		\begin{aligned}
			\,\theta_\mathcal{F}=\{\mathrm{W}^{\mathrm{fi}}_\mathcal{F}, \mathrm{W}^{\mathrm{dr}}_\mathcal{F}\odot\Omega\}\sim q(\theta_\mathcal{F}),\qquad\,\,\,
		\end{aligned}
	\end{equation}
	\begin{equation}
		\begin{aligned}
			\Omega=\{{\mathbf{\omega}_i}\}, \omega\sim\mathrm{Bernoulli}(p_{\mathrm{dropout}}),
		\end{aligned}
		\vspace*{-0.2\baselineskip}
	\end{equation}
	where $\mathrm{W}^{\mathrm{fi}}_\mathcal{F}$ and $\mathrm{W}^{\mathrm{dr}}_\mathcal{F}$
	indicate fixed weights and dropout-involved weights in Forecaster. 
	
	Next, a brief analysis of MC dropout effectiveness is given. Applying the dropout operation in networks is
	mathematically equivalent to a variational approximation of the
	Gaussian process \cite{dropout}, which can be described as follows:
	\begin{equation}		
		\begin{aligned}
			\mathcal{L}_{\mbox{\tiny{GP-MC}}}\propto\xi\sum_{n=1}^{N}\|Y_{i, T_\mathrm{for}:N_T}\!-\!\hat{X}_{i, T_\mathrm{for}:N_T}\|_2^2+\Xi_{p_{\tiny{\mathrm{dropout}}}}\|\mathrm{W}\|_2^2,
		\end{aligned}
	\end{equation}
	where $\mathcal{L}_{\mbox{\tiny{GP-MC}}}$ represents the Formula (20) in the Gaussian process condition with $N$ MC samples and $\xi$ and $\Xi$ denote the constant and varying coefficients, respectively. Therefore, Bayesian approximation can be implemented by using dropout without inferring any additional parameters in ScenGAN.
	
	Given the input samples $X^{\mathrm{enc}}_{\!T:N_T}\!$, $X^{\mathrm{dec}}_{\!T_\mathrm{for}:N_T}\!$, and the latent variable $z$ in the inference stage, the network with dropout $p_{\mathrm{dropout}}$  is fed forward $N_f$ times to obtain the MC samples $\{\hat{X}_{\theta_1}, \hat{X}_{\theta_2}, ..., \hat{X}_{\theta_{\mbox{\tiny{$N_f$}}}}\}$. Then, the epistemic uncertainty can be approximated by the sample variance $\mathrm{\widehat{Var}}(\cdot)$ as:
	\begin{equation}
		\vspace*{-0.2\baselineskip}	
		\begin{aligned}
			\mathrm{\widehat{Var}}(\mathcal{F}_{\theta_\mathcal{F}}\!(X^{\mathrm{enc}}\!, X^{\mathrm{dec}}\!))\!=\!\frac{1}{N_f}\!\!\sum_{n=1}^{N_f}\!\|\hat{X}_{\theta_n}\!\!\!-\!\!\bar{X}\|_1,
		\end{aligned}
	\end{equation}
	where $\bar{X}$ is the average tracks of the MC samples.
	
	[\textit{AdaIN}]: 
	Considering the possible forms of aleatoric uncertainty, the latent variable  $z$ is inserted into the feed-forward network (FFN) of the decoder in a style-based manner\cite{style} as uninterpretable noise for each positional residual, as shown in Fig 3(b); this learnable mapping process is intended to avoid the distributional assumption of stochastic uncertainty. After $N_n$ feed forward procedures with stochastic variable $z$, the sample variance of the deterministic model is computed as:
	\begin{equation}
		\begin{aligned}
			\mathrm{\widetilde{Var}}(\mathcal{F}_{z}(X^\mathrm{enc}\!, X^{\mathrm{dec}}\!))=\frac{1}{N_n}\sum_{n=1}^{N_n}\!\|\hat{X}_{z_n}\!\!-\!\bar{X}\|_1.
		\end{aligned}
	\end{equation}
	\subsubsection{Loss Function}
	As discussed in Subsection II-B, scenario forecasting is viewed as a multimode synthesis process—i.e., it generates multiple scenarios with similar dominant trends—but is different in terms of local characteristics or changes guided by two uncertainties. To infer comprehensive outcomes with limited information, the accuracy and variety of forecasting scenarios are weighed by using a combination of several loss objectives under the adversarial strategy.
	\begin{algorithm*}[b]  
		\caption{: Proposed ScenGAN for Uncertainty-aware Renewable Scenario Forecasting}  
		
		\footnotesize
		\begin{algorithmic}[1] 
			\Require  
			{Number of epoch $N_\mathrm{ep}\!\!=\!\!5000$, learning rate $\alpha\!\!=\!\!0.0008$, batch size $b\!\!=\!\!32$, number of discriminator iterations per Forecaster iteration $n_{\mathrm{d}}\!\!=\!\!2$, numbers of forward propagation per Forecaster iteration $N\!_f\!\!=\!\!8$ and $N\!_n\!\!=\!\!2$. }
			
			\Require
			{Initial weights $\{\theta_\mathcal{F}, \theta_{\mathcal{D}}\}$ for ScenGAN.}  
			\For{epoch $n_\mathrm{ep} \!=\! 0, 1, ..., N_\mathrm{ep}\!-\!1$}
			\For{$n_{\mathrm{d}}$ iterations}
			\State {Sample historical data: $\{X^{\mathrm{hist}}_i\}^{b}_{i=1}	\sim \mathcal{X}_\mathrm{hist}$}
			\State {Sample informational data: $\{X^{\mathrm{info}}_i\}^{b}_{i=1}\!	\sim\! \mathcal{X}_\mathrm{info}$}
			\State {Fetch a batch of input samples according to formula (9) (10) (13):}  \Statex {\qquad\quad$\{\!X^{\mathrm{enc}}_{i, T:N_T}\!\}^{b}_{i=1}$, $\{\!X^{\mathrm{dec}}_{i, T_\mathrm{for}:N_T}\!\}^{b}_{i=1}$, $\{\!X^{\mathrm{dis}}_{i, \hat{T}:N_T}\!\}^{b}_{i=1}$}
			\State {Sample latent variable: $\{z_{i}\}_{i=1}^m \sim \mathcal{Z}$}
			\State{Compute $\mathcal{L} _{D}$ according to formula (30)}
			\State{$\theta_{\mathcal{D}}
				\gets \theta_{\mathcal{D}}-\alpha\cdot \mathrm{AdamOptimizer}(\nabla_{\theta_{\mathcal{D}}}	\mathcal{L} _{D},\theta_{\mathcal{D}}) $}
			\EndFor		
			\State {Sample historical data:$\{X^{\mathrm{hist}}_i\}^{b}_{i=1}	\sim \mathcal{X}_\mathrm{hist}$}
			\State {Sample informational data: $\{X^{\mathrm{info}}_i\}^{b}_{i=1}\!	\sim\! \mathcal{X}_\mathrm{info}$}
			\State {Fetch a batch of input samples according to formula (9) (10):}  
			\Statex {\quad\,\,\,$\{\!X^{\mathrm{enc}}_{i, T:N_T}\!\}^{b}_{i=1}$, $\{\!X^{\mathrm{dec}}_{i, T_\mathrm{for}:N_T}\!\}^{b}_{i=1}$}
			\For{Monte Carlo iteration $n=0, 1, ...,N_{f}\!-\!1$}
			\State {Obtain $\theta_{\mathcal{F}, n}$ according to formula (21)}
			\State {Generate a batch of sequences: $\{\!\hat{X}^{\theta_n}_{i, T_\mathrm{for}:N_T}\!\}^{b}_{i=1}$}
			\EndFor
			\For{Monte Carlo iteration $n=0, 1, ...,N_{n}\!-\!1$}
			\State {Sample latent variable: $\{z_{i, n}\!\}_{i=1}^m \sim \mathcal{Z}$}
			\State {Generate a batch of sequences: $\{\!\hat{X}^{z_n}_{i, T_\mathrm{for}:N_T}\!\}^{b}_{i=1}$}
			\EndFor
			\State{Compute $
				\mathcal{L} _{\mathcal{F}}$
				according to formula (29)}
			\State{$\theta_{\mathcal{F}}
				\gets \theta_{\mathcal{F}}-\alpha\cdot \mathrm{AdamOptimizer}(\nabla_{\theta_{\mathcal{F}}}	\mathcal{L} _{\mathcal{F}},\theta_{\mathcal{F}}) $}
			
			\EndFor
			
		\end{algorithmic}
	\end{algorithm*}
	
	[\textit{Variety Loss}]:   
	First, a variety loss term is introduced to encourage Forecaster to generate diverse scenarios covering epistemic uncertainty. For each iteration, $N_f$ outputs are randomly sampled through the repeated forward-propagation procedures, and the optimal output in the sense of the Euclidean norm is chosen to participate in the optimization procedure:
	\begin{equation}		
		\begin{aligned}
			\mathcal{L}^{\mathrm{va}}_{\mathcal{F}}=\min_{N_f}\parallel\! Y_{T_{\mathrm{for}}:N_T}\!-\! \hat{X}^{\theta_{N_f}}_{T_{\mathrm{for}}:N_T}\parallel_2^2,
		\end{aligned}
	\end{equation}
	where $Y_{T_{\mathrm{for}}:N_T}$ represents ground truth. The method of specifying the optimal sample is matched with Bayesian inference to cover the output space of future scenarios.
	
	[\textit{Auxiliary Loss}]: Then, an auxiliary loss term that decays during the training iteration procedure is added to maintain the inchoate forecasting distribution calibration as follows:
	\begin{equation}
		\begin{aligned}
			\mathcal{L}^{\mathrm{au}}_{\mathcal{F}}=e^{\!\!-\epsilon n}\mathbb{E}_{_{z\sim\mathcal{Z}}^{\theta\sim\Theta,}}\!\big[\big\|\! Y_{T_{\mathrm{for}}:N_T}\!\!-\!\! \mathcal{F}_{\theta}(X_{\!T:N_T}^\mathrm{enc}, X_{T_{\mathrm{for}}:N_T}^\mathrm{dec}, z)\big\|_2^2\big],
		\end{aligned}
	\end{equation}
	where $e^{\!\!-\epsilon n}$ is the exponential decay term with index $\epsilon$ and $n$ indicates the current order of iterations. 
	
	[\textit{Adversarial Loss}]: As a kind of generator, Forecaster $\mathcal{F}$'s optimization also involves an adversarial strategy, which considers the outputs of the discriminator $\mathcal{D}$, and identifies the renewable authenticity of the generated forecasts:
	\begin{equation}
		\begin{aligned}
			\mathcal{L}^{\mathrm{ad}}_{\mathcal{F}}=\mathbb{E}_{_{z\sim\mathcal{Z}}^{\theta\sim\Theta,}}\big[-\mathcal{D}(\mathcal{F}_{\theta}(X_{\!T:N_T}^\mathrm{enc}, X_{T_{\mathrm{for}}:N_T}^\mathrm{dec}, z))\big].
		\end{aligned}
	\end{equation}
	
	Overall, the Forecaster loss can be summarized as follows:
	\begin{equation}
		\begin{aligned}
			\mathcal{L}_{\mathcal{F}}=\lambda_{\mathrm{va}}\!\mathcal{L}^{\mathrm{va}}_{\mathcal{F}} + \lambda_{\mathrm{au}}\!\mathcal{L}^{\mathrm{au}}_{\mathcal{F}} + \lambda_{\mathrm{ad}}\!\mathcal{L}^{\mathrm{ad}}_{\mathcal{F}} + 
			\widetilde{\sigma}(N_n), 
		\end{aligned}
	\end{equation}
	where $\lambda_{\!\mathrm{va}}$, $\lambda_{\!\mathrm{au}}$ and $\lambda_{\!\mathrm{ad}}$ indicate the contribution coefficients, and $\widetilde{\sigma}(N_n)=\frac{1}{N_n}\!\!\sum_{n=1}^{N_n}\!\!\|Y_{T_{\mathrm{for}}:N_T}\!-\!\hat{X}^{z_n}_{T_{\mathrm{for}}:N_T}\|_2^2$ acts as a remedy term to address noise.
	
	[\textit{Hinge Loss}]: For discriminator, a hinge loss $\mathcal{L_D}$ designed for
	criteria construction is implemented with a gradient penalty, which helps stabilize the optimization procedure of the discriminator and improve the authenticity of the scenarios:
	\begin{equation}
		\begin{aligned}
			\mathcal{L}_{\mathcal{D}}^{\mathrm{+}}=\mathbb{E}_{_{z\sim\mathcal{Z}}^{\theta\sim\Theta,}}\big[\max(0, 1+\mathcal{D}(\mathcal{F}_{\theta}(X_{\!T:N_T}^\mathrm{enc}, X_{T_{\mathrm{for}}:N_T}^\mathrm{dec}, z)))\big],\,\,
		\end{aligned}
	\end{equation}
	\begin{equation}
		\begin{aligned}
			\mathcal{L}_{\mathcal{D}}^{\mathrm{-}}=\mathbb{E}_{X^{\mathrm{dis}}_{\hat{T}:N_T}\!\!\sim\mathcal{X}_\mathrm{hist}}\big[\max(0, 1-\mathcal{D}(X_{\hat{T}:N_T}^{\mathrm{dis}}))\big],\qquad\qquad\,\,\,
		\end{aligned}
	\end{equation}
	\begin{equation}
		\begin{aligned}
			\mathcal{L}_{\mathcal{D}}=\mathcal{L}_{\mathcal{D}}^{\mathrm{+}}+\mathcal{L}_{\mathcal{D}}^{\mathrm{-}}+\lambda_{\mathrm{gp}}\mathbb{E}_{X^{\mathrm{dis}}_{\hat{T}:N_T}\!\!\!\sim\mathcal{X}_\mathrm{hist}}\big[\|\nabla_\theta\mathcal{D}(X_{\hat{T}:N_T}^{\mathrm{dis}})\|_2\big],\,\,\,
		\end{aligned}
	\end{equation}
	where $\lambda_{\mathrm{gp}}$ is the coefficient of R1 regularization term \cite{training}. The training procedure of ScenGAN using these designed losses is shown in Algorithm 1.
	\section{Case Studies}
	In this section, the datasets and model configurations are first described, and the forecasting results are subsequently presented. Then, ablation studies and contrastive analysis are evaluated on different evaluation metrics. Finally, stochastic optimization for unit commitment is performed to test the day-ahead scenario effectiveness.
	\subsection{Data Specification and Model Setting}
	To comprehensively verify the feasibility of applications, ScenGAN is trained on three data sources. In the first case, we select historical power records published by Elia Group \cite{elia}, a Belgian transmission system operator, as a regional aggregation-level data source. In addition, the meteorological datasets, which were obtained from the Wunderground Database \cite{elia-weather} and processed by the nearest interpolation, are employed as independent meteorological input sources in the first case to further increase the dimension of information. For comparison, the second case focuses on multivariable power sequences provided by the National Renewable Energy Laboratory (NREL) \cite{nrel}. In the third case, we use datasets from GEFCom2014 \cite{GEFCOM} that are freely accessible for reproducible studies to train models and conduct experiments on the stochastic optimization of the day-ahead unit commitment model. More details about these datasets can be found in Table 2. To facilitate interpretation, the power values of the three cases are normalized according to the installed capacities of their electric fields. 
	\begin{table*}[h]
		\caption{Details of the Evaluated Data Sets}
		\renewcommand{\arraystretch}{1.}
		\begin{center}  	
			\setlength{\tabcolsep}{4mm}{
				\begin{tabular}{cccc}
					\toprule[0.5mm]
					&Case $\mathbf{A}$ &Case $\mathbf{B}$ &Case $\mathbf{C}$\\
					\midrule
					Data Source & Elia, Wunderground & NREL &  GEFCom2014 \\
					\midrule
					\multirow{3}{*}{Description}&Regional  wind \& PV, &20 wind farms, &Regional wind \& PV, \\& meteorology, &  30 PV plants & meteorology\\ & multiscale forecasts &  &   \\
					\midrule
					Location & Offshore Belgium & WA., USA & Australia\\
					\midrule
					Resolution&15 minutes& 5 minutes & 60 minutes\\
					\midrule
					Effective&Wind:  $2020/2/1$---$ 2022/2/28$& Wind:  $2007/1/1$---$ 2013/12/31$&Wind:  $2012/1/1$---$2012/10/1$\\ 
				
					Period	&PV: $2020/3/31$---$2022/7/31$ & PV: $2006/1/1$---$2006/12/31$&PV: $2012/4/1$---$2013/5/1$\\
				
					\midrule
					\multirow{2}{*}{Lead Time} & Day-ahead, & Day-ahead, & Week-ahead,\\ &96 nodes &288 nodes & 168 nodes \\
					\midrule
					$\mathcal{X}_{\mathrm{hist}}$& \multicolumn{3}{c}{Overall historical power tracks}\\
					\midrule
					\multirow{2}{*}{$\mathcal{X}_{\mathrm{info}}$}& Point forecast, & Historical power& NWP\\& NWP & & \\
					\bottomrule[0.5mm]				
			\end{tabular}}
			\label{tab1}
		\end{center}
	\end{table*} 	
	
	%The sets $\mathcal{X}_{\mathrm{hist}}$ and $\mathcal{X}_{\mathrm{info}}$ in each case is also listed in Table 2.
	
	In each case, a standard division ratio of 4:1:1 is adopted for constructing the training set, validation set and testing set. Considering the task complexity and computational efficiency, a random search method is implemented for adjusting the hyperparameters (except the fixed $N_T$, $\widetilde{N}_T$ and $N_m$), during which the parameter set tuning path is controlled by the accuracy score obtained on the validation set. The tuning results are given in Table 3. For comparison, 10 benchmarks, including 1 modified ARIMA method, 6 DGM-based scenario generation models, and 3 sequence-based networks, are deployed in the evaluation as follows: 
	\begin{itemize}
		\item SLARIMA \cite{slarima}:  
		Seasonal ARIMA model with a limiter.
		\item DeepAR \cite{deepar}:  
		LSTM-based probabilistic model.
		\item DCGAN-SO \cite{s2}: Deep convolutional GAN model.
		\item StyleGAN-SO:  
		Modified version of \cite{s2} in a style-based manner.
		\item SeqGAN-LSTM \cite{s12}: Sequence GAN model.
		
		\item ProGAN-MO \cite{s13}:  
		Progressive growing GAN model.
		\item StyleGAN2-SE \cite{s16}:  
		Modified style-based GAN model.
		\item MLSTM: 4-layer multiLSTM model. 
		\item Transformer \cite{transformer}:  
		Sequence-to-Sequence attention-based model.
		\item Informer \cite{informer}:  
		Variant version of the transformer model.  
	\end{itemize}
	\begin{table*}[h]
		\caption{Results of Parameter Tuning}
		\renewcommand\arraystretch{1}
		\begin{center}
			\setlength{\tabcolsep}{15mm}{
				\begin{tabular}{ccc}
					\toprule[0.5mm]
					Notation & Parameter Description &Value ($\mathbf{A} / \mathbf{B} / \mathbf{C}$) \\
					\midrule
					$N_T$ & Number of whole length & 192  / 576 / 288 \\
					$\widetilde{N}_T$ &Number of lagging length& 96 / 288/ 120\\
					$N_m$ &Number of feature dimension& 100 / 100 / 100\\
					$N_\mathrm{e}$ &Number of distilling modules& 2 / 2 / 2\\
					$N_f$ &Number of feature patterns& 8 / 8 / 8\\
					$N_n$ &Number of latent variable sampling& 2 / 2 / 2\\
					$h$ &Number of attention heads& 8 / 8 / 8\\
					$p_{\mbox{\tiny{dropout}}}$ &Ratio of MC dropout& 0.2 / 0.2 / 0.2\\
					$\epsilon$ &Index of decay term& 0.05 / 0.05 / 0.05\\
					$\lambda_{\mathrm{gp}}$ &Coefficient of R1 regularization& 3.0 / 2.0 / 3.0\\
					$[\lambda_\mathrm{va}, \lambda_{\mathrm{au}}, \lambda_{\mathrm{ad}}]$ &Coefficients of Forecaster loss& [0.5, 0.5, 1.0]\\
					\bottomrule[0.5mm]	
			\end{tabular}}
			\label{tab1}
		\end{center}
	\end{table*}
	
	All computations were deployed on a cloud server running the Ubuntu 16.04.7 LTS operating system with 48 Intel Xeon E5-2650 v4 CPUs, 10 PH402 SKU 200 GPUs and a total of 188 GB RAM. The model was implemented using Python 3.7.7 with TensorFlow 2.1.0 and Keras 2.3.1. The testing simulation was constructed with GurobiPy 9.5.0.
	\subsection{Case $\mathbf{A}$: Interpretability Description}
	First, we describe how ScenGAN performs forecasts with the attention mechanism and explain the representations of uncertainties. For case $\mathbf{A}$, we organize the informational sample into a matrix combining multiscale point forecasts and meteorological time series:
	\begin{equation}
		\begin{aligned}
			X_{\mathbf{A}}^{\mathrm{info}}=[\mathring{\boldsymbol{x}}^{\mathbf{for1}}, \mathring{\boldsymbol{x}}^{\mathbf{for2}}, \mathring{\boldsymbol{x}}^{\mathbf{for3}}, \mathring{\boldsymbol{c}}^{\mathbf{t}}, 
			\mathring{\boldsymbol{c}}^{\mathbf{d}},
			\mathring{\boldsymbol{c}}^{\mathbf{h}},
			\mathring{\boldsymbol{c}}^{\mathbf{w}},	
			\mathring{\boldsymbol{c}}^{\mathbf{p}}	
			]^\intercal,
		\end{aligned}
	\end{equation}
	\begin{equation}
		\begin{aligned}
			\mathring{\boldsymbol{c}}^{\bullet}=\mathrm{concat}(\boldsymbol{c}^\mathbf{hist}_{T:\widetilde{N}_T}, \boldsymbol{c}^\mathbf{nwp}_{T_{\mathrm{for}}:N_T-\widetilde{N}_T}),\qquad\quad\,
		\end{aligned}
	\end{equation}
	where $\{\mathring{\boldsymbol{x}}^{\mathbf{for1}}, \mathring{\boldsymbol{x}}^{\mathbf{for2}}, \mathring{\boldsymbol{x}}^{\mathbf{for3}}\}$ represents the point forecasts in the range of a week to a day ahead;  $\{\mathring{\boldsymbol{c}}^{\mathbf{t}}, 
	\mathring{\boldsymbol{c}}^{\mathbf{d}},
	\mathring{\boldsymbol{c}}^{\mathbf{h}},
	\mathring{\boldsymbol{c}}^{\mathbf{w}},	
	\mathring{\boldsymbol{c}}^{\mathbf{p}}\}$ indicates weather information sequences, including temperature (t), dew point (d), humidity (h), wind speed (w), and pressure (p); and $\{\boldsymbol{c}^\mathbf{hist}, \boldsymbol{c}^\mathbf{nwp}\}$ represents historical observations and day-ahead NWP.
	
	\subsubsection{Uncertainty Representation}
	
	We represent renewable uncertainties in a ``hierarchical" way by using the Bayesian deep model to describe possible patterns inferred by MC dropout and incorporating latent variables to express random mutations. We show these two levels of uncertainty in Fig. 4 and Fig. 5, respectively. In Fig. 4, different pattern outputs controlled by model parameters are used to infer the trends of renewable development trajectories while retaining certain local diversity and randomness. In addition, model parameter randomization can improve the coverage ability of the ground truth for renewable power forecasting. In Fig. 5, scenarios in the same pattern inherit the overall curve trend, power generation level and local description from the model, providing a diversity scheme with only subtle variations. This can be seen as a promising property for renewable scenario forecasting, as the proposed model is able to analyze and capture epistemic uncertainty and disentangle renewable uncertainties at different time-varying levels.
	\begin{figure*}[h]	
		
		\centerline{\includegraphics[width=18.0cm]{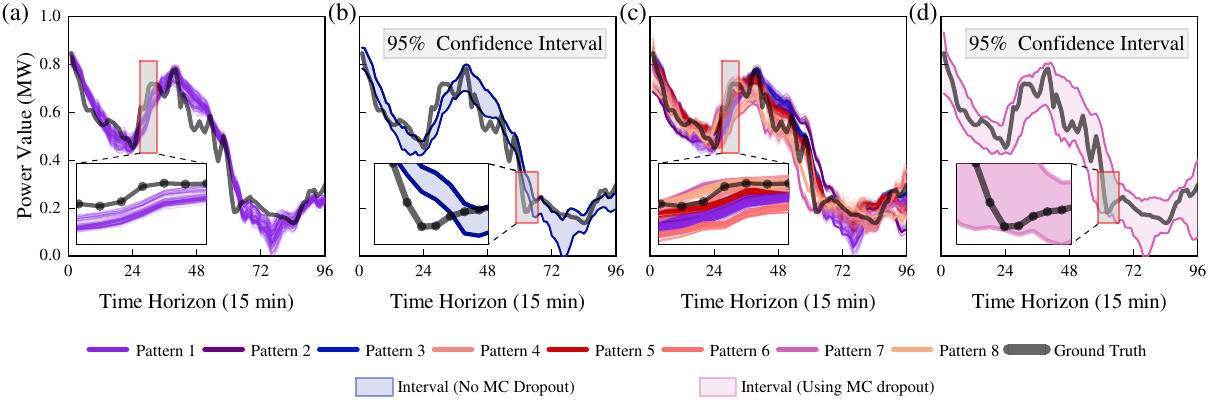}}
		\caption{Performance of wind scenario forecsting before (a) (b) and after (c) (d) epistemic uncertainty awareness.\\}
		\label{fig}	
		\centerline{\includegraphics[width=18.0cm]{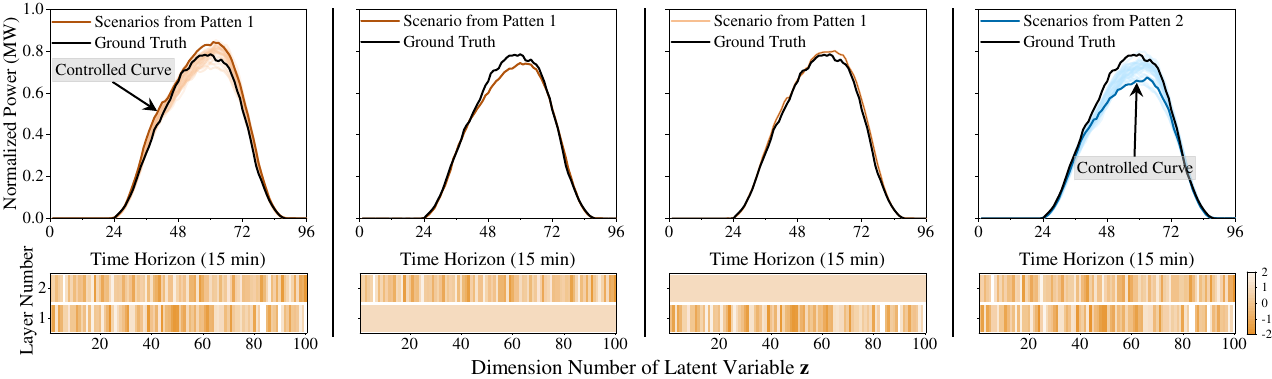}}
		\caption{Impact of latent variables on PV scenario generation}
		\label{fig}
	\end{figure*}
	
	\subsubsection{Mechanism of Temporal Attention}
	
	To provide persuasive insights, Fig. 6 shows the scenario forecasting results tested from Jan. 30 to Feb. 1, 2022 and the Softmax scores given by the last cross-attention layer of Forecaster's decoder block. It can be seen that, even under conditions where multiscale point forecasts are inaccurate, single-forward ScenGAN is able to generate power behaviors that are relatively consistent with real observations, largely due to the capture of complex patterns by attention modules. In Fig. 6(b), the Softmax score gives higher weights to fast-ramp, high-peak and local-fluctuation segments, further indicating that the trained network can accurately grasp the key information for the future.
	\begin{figure*}[h]	

		\centerline{\includegraphics[width=18.0cm]{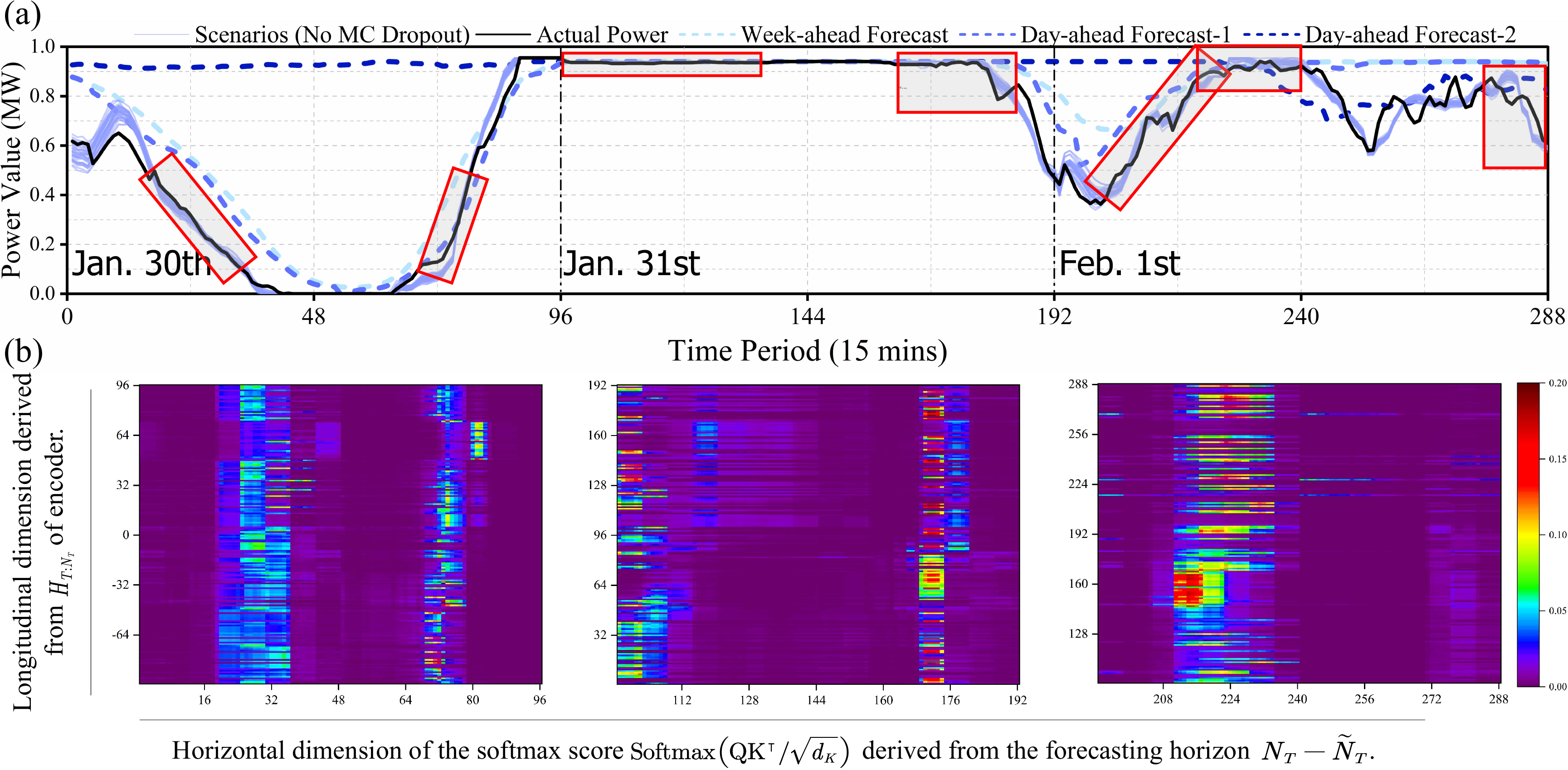}}

		\caption{Inside view of scenario forecasting results for days from Jan 30th to Feb 1st, 2022. Wind scenarios (no MC Dropout) (a). Softmax scores in the last cross-attention module trained on Elia (b).}	
	\end{figure*}
	
	\begin{figure*}[h]

		\centerline{\includegraphics[width=18.0cm]{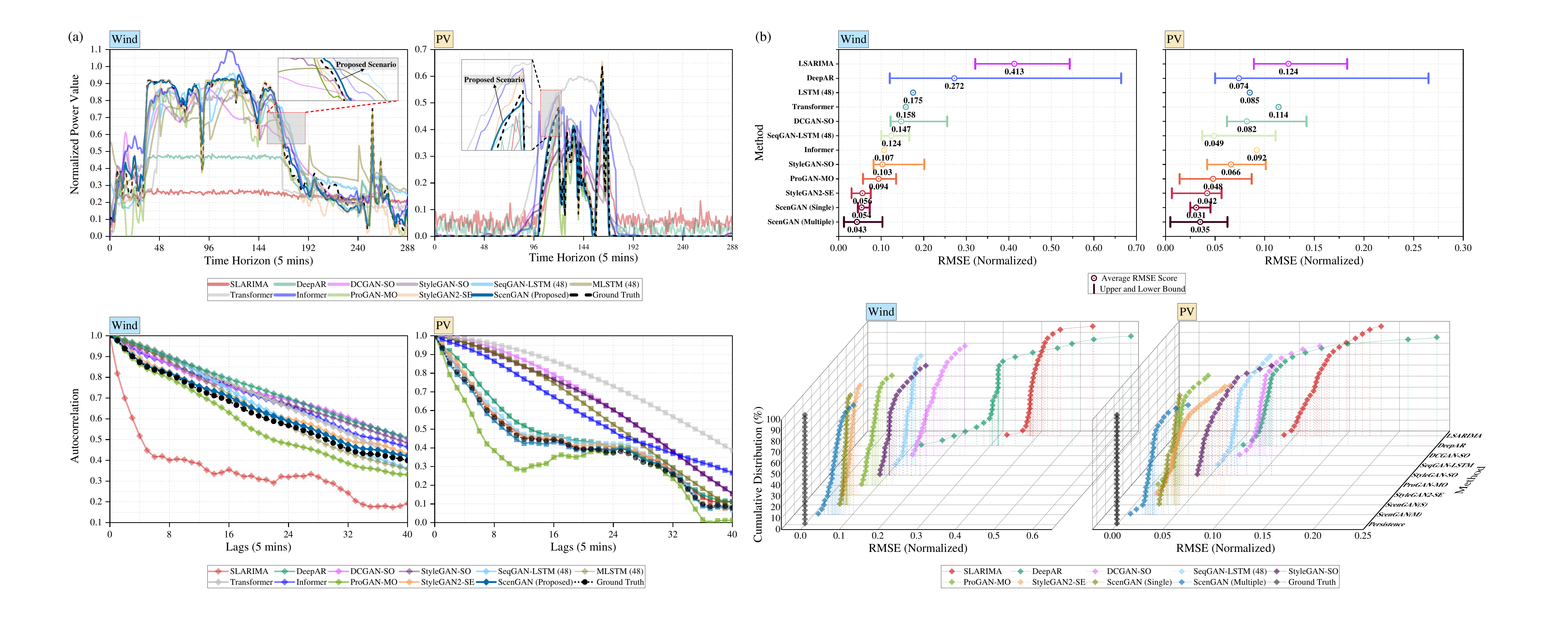}}
		\caption{Comparison of day-ahead forecasts using various models for wind power of May 22nd, 2013 and PV power of Dec. 31st, 2006. Forecasted trajectories and autocorrelation diagrams (a). RMSE interval and cumulative distribution diagrams of forecasts generated by various models (b).}
	\end{figure*}
	
	\subsection{ Case $\mathbf{B}$: Comparison and Ablation Experiments }
	Based on the NREL dataset, comparison and ablation experiments are designed to verify the improvement of ScenGAN in renewable scenario forecasting and the effectiveness of the employed modules. For fairness, all models are trained in the same data conditions; the informational input of ScenGAN is defined as $X_{\mathbf{B}}^{\mathrm{info}}=[\mathbf{x}^{(1)}, \mathbf{x}^{(2)}, 
	...,	
	\mathbf{x}^{(N_\rho)}	
	]^\intercal$,
	where $\mathbf{x}^{(\bullet)}$
	represents the historical tracks of Washington area sites, and $N_\rho$ takes the value 30 (PV) and 20 (Wind), respectively.
	\begin{sidewaystable*}[htbp]
		\caption{Comparative Experiments of Different Methods in Case $\mathbf{B}$}
		\renewcommand\arraystretch{1}
		\begin{center}
			\linewidth
			\setlength{\tabcolsep}{}{
				\begin{tabular}{ccccccccccccccccc}
					\toprule[0.5mm]
					\multirow{3}{*}{Method} &\multicolumn{8}{c}{Wind Power}& \multicolumn{8}{c}{PV Power}\\
					\cmidrule(lr){2-9}\cmidrule(lr){10-17}
					&\multicolumn{2}{c}{Deterministic }&\multicolumn{6}{c}{Probabilistic}&\multicolumn{2}{c}{Deterministic}&\multicolumn{6}{c}{Probabilistic}\\
					\cmidrule(lr){2-3}\cmidrule(lr){4-9}\cmidrule(lr){10-11}\cmidrule(lr){12-17}
					&$\overline{\mbox{RMSE}}$&$\mbox{S}_{\mathrm{score}}$&$\overline{\mbox{PS}}$&$\overline{\mbox{BS}}$&$\overline{\mbox{CRPS}}$&$\mbox{SS}_{\overline{\mbox{\tiny{CRPS}}}}$&ES&WS&$\overline{\mbox{RMSE}}$&$\mbox{S}_{\mathrm{score}}$&$\overline{\mbox{PS}}$&$\overline{\mbox{BS}}$&$\overline{\mbox{CRPS}}$&$\mbox{SS}_{\overline{\mbox{\tiny{CRPS}}}}$&ES&WS\\
					
					\midrule
					SLARIMA &0.3435& 0.6994 & 0.0907 &0.4769&0.3164&-0.8744&2.3061&1.4367&0.1824& 0.2377 &0.0354&0.1701&0.1032&-0.6618&1.1517&0.2651\\
					DeepAR & 0.3017 & 0.6940 & 0.0713 &0.4312&0.2555&-0.5136&2.0475&1.1389&0.1685& 0.1879 &0.0286&0.1637&0.0856&-0.3784&1.1377&0.2445\\
					DCGAN-SO & 0.2593 & 0.2947 & 0.0689  &0.3285&0.1679&0.0053& 1.3929 &1.2382&0.1195&0.1224&0.0241&0.1400&0.0507&0.1836&0.9231&0.1686\\
					StyleGAN-SO & 0.1698 & 0.1889 & 0.0411 &0.2222&0.1157&0.3146&1.1451&0.8283&0.0717&0.1244&0.0169&0.1339&0.0328&0.4718&0.7101 &0.1276\\
					SeqGAN-LSTM & 0.1642 & 0.1728 & 0.0323 &0.1335&0.1101&0.3477&1.0583&0.9258&0.0585&0.0598&0.0086&0.1095&0.0267&0.5700&0.6621&0.1382\\
					MLSTM & 0.2274 & 0.5315 &-&-&-&-&-&-&0.1040&0.1661&-&-&-&-&-&-\\
					Transformer& 0.2551 & 0.5217 &-&-&-&-&-&-&0.1218& 0.1845 &-&-&-&-&-&-\\
					Informer& 0.1318 & 0.2881 &-&-&-&-&-&-&0.1009&0.1483&-&-&-&-&-&-\\
					ProGAN-MO & 0.1169 & 0.1258 & 0.0204 &0.1235&0.0706&0.5818&0.9677&0.6930&0.0614&0.0655&0.0093&0.1008&0.0183&0.7053&0.6558&0.0898\\
					StyleGAN2-SE& 0.0935&0.0897&0.0133&0.1068&0.0534&0.6836&0.8823&0.4426&0.0561&0.0530&0.0056&0.0990&0.0151&0.7568&0.4893&0.0799\\
					\midrule
					ScenGAN$^\dag$	&0.0851 & 0.0836 &0.0152&\cellcolor{Aquamarine1}\textbf{0.0838}&0.0541&0.6795&0.9027&0.4519&0.0472& 0.0459 &0.0076&0.0751&0.0198&0.6812&0.4481&0.1077\\
					ScenGAN  & \cellcolor{Aquamarine1}\textbf{0.0814} &\cellcolor{Aquamarine1}\textbf{0.0805} &\cellcolor{Aquamarine1}\textbf{0.0102}&0.1008&\cellcolor{Aquamarine1}\textbf{0.0465}&\cellcolor{Aquamarine1}\textbf{0.7245}&\cellcolor{Aquamarine1}\textbf{0.7762}&\cellcolor{Aquamarine1}\textbf{0.3771}&\cellcolor{Aquamarine1}\textbf{0.0445}& \cellcolor{Aquamarine1}\textbf{0.0438} & \cellcolor{Aquamarine1}\textbf{0.0047} &\cellcolor{Aquamarine1}\textbf{0.0686}&\cellcolor{Aquamarine1}\textbf{0.0131}&\cellcolor{Aquamarine1}\textbf{0.7890}&\cellcolor{Aquamarine1}\textbf{0.3239}&\cellcolor{Aquamarine1}\textbf{0.0591}\\
					
					\bottomrule[0.5mm]	
					
			\end{tabular}}
			\label{tab1}
		\end{center}
		\textbf{ScenGAN$^{\dag}$}: deterministic ScenGAN without MC dropout.
		
	\end{sidewaystable*} 
	\subsubsection{Comparison}
	To verify the superiority of the proposed method, day-ahead forecast is performed for wind power on May 22nd, 2013, and for PV power on Dec. 31st, 2006, based on the aforementioned 10 trained models and ScenGAN. Fig. 7(a) exhibits the visual inspection results of the forecasts and the corresponding time-series autocorrelation with varied time lags. For the probabilistic forecasting models, we utilize the average curves in the comparison process. Compared with those of the other models, the expected forecast inferred by ScenGAN accords best with the actual power behavior in terms of trends and time-domain correlations, which is largely due to its grasp of the multilevel temporal structure, especially the local patterns, in the forecasting process. Then, we incorporate renewable uncertainties and investigate the forecasting performance comparison among the various models through the distribution of the root-mean-square error (RMSE). In this process, ScenGAN is separated into two parts that represent certain and uncertain parameters, as shown in Fig. 7(b). The RMSE distribution diagrams show that the RMSE distributions from the LSARIMA to the ScenGAN models present a progressive downward trend, which is attributed to ScenGAN’s consideration of forecasting authenticity and pattern diversity. In addition, although ScenGAN with uncertain parameters has a greater uncertainty degree than deterministic ScenGAN, uncertain ScenGAN can infer more accurate patterns, and its overall performance is more reliable.
	
	Furthermore, randomly chosen 30-consecutive-days ($N_\mathrm{d}\!\!\!=\!\!\!30$) forecasts are performed to quantitatively assess the deterministic and probabilistic qualities of the methods, where the scenario forecasting models participate in the assessment for 100 generated scenarios ($N_s\!\!=\!\!100$).
	
	Given the actual observations and forecasted results, two deterministic statistical metrics, average RMSE ($\overline{\mbox{RMSE}}$) and a defined score ($	\mathrm{S}_\mathrm{score}$) proposed in \cite{kddcup} are calculated to assess the average performance of the different benchmarks. To evaluate the performance of renewable
	probabilistic forecasting methods, six well-known metrics are applied in this case to measure the consistency, variability, and
	tightness of the estimated distributions, which are pinball score
	($\overline{\mbox{PS}}$), Brier score ($\overline{\mbox{BS}}$) \cite{s13}, energy score (ES), Winkler score (WS) \cite{bayesian4}, average continuous ranked probability score ($\overline{\mbox{CRPS}}$), and  $\overline{\mbox{CRPS}}$ skill score ($\mathrm{SS}_{\overline{\mbox{\tiny{CRPS}}}}$). Thereinto, $\overline{\mbox{CRPS}}$ proposed in this
	study is expressed as the difference between the forecasted and
	the persistent distributions, and $\mathrm{SS}_{\overline{\mbox{\tiny{CRPS}}}}$ compares the results with the official NWP-based point forecasts to determine the degree of improvement of each benchmark intuitively:
	\begin{equation}
		\overline{\mbox{CRPS}} = \frac{1}{288N_{\mathrm{d}}}\sum^{N_{\mathrm{d}}}_{i=1}\sum_{t=0}^{288}\int_{0}^{1}(\hat{F}_{i,t}(\hat{y})-\mathbf{1}(\hat{y}\geq y_{i,t}))d\hat{y},
	\end{equation}
	\begin{equation}
		\,\,\,\,\,\,\,\,\,\mathrm{SS}_{\overline{\mbox{\tiny{CRPS}}}} = 1 - {\overline{\mbox{CRPS}}}/{\overline{\mbox{MAE}}_{\mbox{\scriptsize{pf}}}},\qquad\qquad\qquad\qquad\qquad
	\end{equation}
	where $\hat{F}(\cdot)$ represents the cumulative distribution function, and $\overline{\mbox{MAE}}_{\mbox{\scriptsize{pf}}}$ is the mean absolute error of the provided point forecasts. According to the results summarized in Table 4, with the best scores in bold, the single-forward ScenGAN and normal ScenGAN outperform the others on the deterministic and probabilistic metrics, respectively, because the deterministic ScenGAN’s output is more in line with the most likely characteristics of RES generation, while the uncertain ScenGAN is more focused on diversifying expression. It is worth noting that StyleGAN2-SE also achieves competitive scores on both aspects, largely due to its style-based scenario generation mode and efficient forecasting networks.
	\subsubsection{Ablation Experiment}
	To verify the rationality of the various modules and techniques mentioned in this study, ablation experiments are designed to track the changes in sharpness, reliability and spatial-temporal correlation characteristics. In addition to the above two models, six ScenGAN counterparts are designed in this process: i) Forecaster: employ Forecaster with
	MSE loss naively. ii) Forecaster$^\ast$: remove the Probparse attention
	and distilling modules from Forecaster. iii) Forecaster$^\ddagger$
	: remove the multilevel embedding from Forecaster. iv) ScenGAN$^\ddagger$: employ ScenGAN, but remove its multilevel embedding.
	v) ScenGAN$^{\mbox{\tiny{msd}}}$: remove the mini-batch standard deviation module from ScenGAN. vi) ScenGAN$^{\blacklozenge}$: remove the spatial-based
	convolution module from ScenGAN. In addition, the transparent
	exploration process for the parameter $p_{\mathrm{dropout}}$ demonstrates its impact on forecasting effectiveness. Note that, a modified variogram score $\mathrm{V}_\mathrm{score}^k$ is employed to capture spatial correlations:
	\begin{equation}
		\mathrm{V}_\mathrm{score}^k =\sum_{i=1}^{N_\rho}\sum_{j=1}^{N_\rho}\big|\|Y^{(i)}\!-\!Y^{(j)}\|_k-\frac{1}{N_s}\sum_{n=1}^{N_s}\|\hat{Y}_n^{(i)}\!-\!\hat{Y}_n^{(j)}\|_k\big|,
	\end{equation}
	where $Y$ and $\hat{Y}$ denote the actual sequence and scenario at a
	certain power site. In addition, the average proportion deviation
	(APD) proposed by \cite{distribution} is extended to the reliability measure of renewable scenario forecasting, as shown in Table 5. According to the statistical results, the removal of the Probparse self-attention modules and temporal embedding decreases forecasting quality, demonstrating the effectiveness and rationality of the designed temporal embedding technique and attention structure. The GAN-based architecture and mini-batch standard deviation strategy improve the capacity of uncertainty description, making the pattern expression more diverse. The measurement results of spatial correlation also demonstrate the improvement of spatial perception by the spatial-based convolution module in the multivariate condition. In addition, the proposed method exhibits acceptable forecasting performance under different dropout probabilities, indicating competitive robustness under erratic environments.
	\begin{sidewaystable*}[htp]
		\caption{Ablation Study under Spatial-temporal Correlation}
		\renewcommand\arraystretch{1}
		\begin{center}
		\setlength{\tabcolsep}{3.6mm}{
				\begin{tabular}{ccccccccccccc}
					\toprule[0.5mm]
					\multirow{3}{*}{Backbone} &\multicolumn{6}{c}{Wind Power}& \multicolumn{6}{c}{PV Power}\\
					\cmidrule(lr){2-7}\cmidrule(lr){8-13}
					&\multicolumn{2}{c}{Reliability} & \multicolumn{2}{c}{Sharpness}  &  \multicolumn{2}{c}{Space-time}&\multicolumn{2}{c}{Reliability} & \multicolumn{2}{c}{Sharpness}  &  \multicolumn{2}{c}{Space-time}\\	\cmidrule(lr){2-3}\cmidrule(lr){4-5}\cmidrule(lr){6-7}\cmidrule(lr){8-9}\cmidrule(lr){10-11}\cmidrule(lr){12-13}
					&$\mathrm{S}_\mathrm{score}$&APD (\%)&$\mathrm{\widehat{Var}}$&$\mathrm{\widetilde{Var}}$&$\mathrm{V}_\mathrm{score}^1$&$\mathrm{V}_\mathrm{score}^2$&$\mathrm{S}_\mathrm{score}$&APD (\%)&$\mathrm{\widehat{Var}}$&$\mathrm{\widetilde{Var}}$&$\mathrm{V}_\mathrm{score}^1$&$\mathrm{V}_\mathrm{score}^2$\\
					\midrule
					Forecaster$^\ast$&0.1441& 6.47 & - &0.0474&11.6911&1.6423&0.0775&11.26&-&\cellcolor{Aquamarine1}\textbf{0.0352}&6.4703&0.9193\\
					Forecaster$^\ddagger$&0.1572& 7.53 & - &0.0605&10.7489&1.3445&0.0818&18.91&-&0.0414&6.0194&0.8241\\
					Forecaster&0.0976&4.98&-&\cellcolor{Aquamarine1}\textbf{0.0417}&10.0886&1.0083&0.0649&4.62&-&0.0433&5.3381&0.7109\\
					ScenGAN$^\dag$&\cellcolor{Aquamarine1}\textbf{0.0805}&1.89&-&0.1415&9.4831&0.8163&\cellcolor{Aquamarine1}\textbf{0.0438}&4.37&-&0.1041&5.0633&0.6249\\
					ScenGAN$^\ddagger$&0.1280&5.30&0.1025&0.0483&9.7439&0.9064&0.0756&9.37&0.0453&0.0590&5.2215&0.6551\\
					ScenGAN$^{\mbox{\tiny{msd}}}$&0.0899&3.41&0.0639&0.0781&8.9030&0.7938&0.0501&4.79&0.0381&0.0658&4.7485&0.4831\\
					ScenGAN$^{\blacklozenge}$&0.0870&2.95&0.0811&0.0627&10.4832&1.2530&0.0522&5.24&0.0454&0.0512&6.2126&0.8746\\
					\midrule
					\multicolumn{13}{c}{ScenGAN}\\
					\midrule
					$p_{\mathrm{dropout}}=0.50$&0.1105&1.77&0.0901&0.0535&9.0397&0.8147&0.0662&4.88&0.0563&0.0489&4.9908&0.4176\\
					$p_{\mathrm{dropout}}=0.40$&0.0994&1.35&0.0792&0.0642&8.7232&0.7227&0.0563&4.13&0.0445&0.0528&4.4437&0.3238\\
					$p_{\mathrm{dropout}}=0.30$&0.0918&1.06&0.0648&0.0710&8.1855&0.5736&0.0484&3.89&0.0321&0.0662&4.0058&0.2309\\
					$p_{\mathrm{dropout}}=0.20$&0.0851&$\textbf{0.82}$&0.0445&0.0743&\cellcolor{Aquamarine1}\textbf{7.5687}&\cellcolor{Aquamarine1}\textbf{0.3467}&0.0459&\cellcolor{Aquamarine1}\textbf{3.70}&0.0209&0.0708&\cellcolor{Aquamarine1}\textbf{3.5123}&\cellcolor{Aquamarine1}\textbf{0.1409}\\
					$p_{\tiny\mathrm{dropout}}=0.05$&0.0842&1.25&\cellcolor{Aquamarine1}\textbf{0.0372}&0.0906&8.2965&0.6016&0.0451&4.24&\cellcolor{Aquamarine1}\textbf{0.0126}&0.0821&3.9203&0.1924\\
					\bottomrule[0.5mm]	
					
			\end{tabular}}
			\label{tab1}
		\end{center}
		\textbf{Forecaster$^\ast$}: a canonical attention version of Forecaster.  \textbf{Forecaster$^\ddagger$}: remove multilevel embedding in Forecaster. \textbf{ScenGAN$^\ddagger$}: remove multilevel embedding in ScenGAN.
		\textbf{ScenGAN$^{\mbox{\tiny{msd}}}$}: remove the mini-batch standard deviation module.
		\textbf{ScenGAN$^{\blacklozenge}$}: remove the spatial-based convolution module.
	\end{sidewaystable*} 
	\begin{figure*}[htbp]
		\centerline{\includegraphics[width=18.0cm]{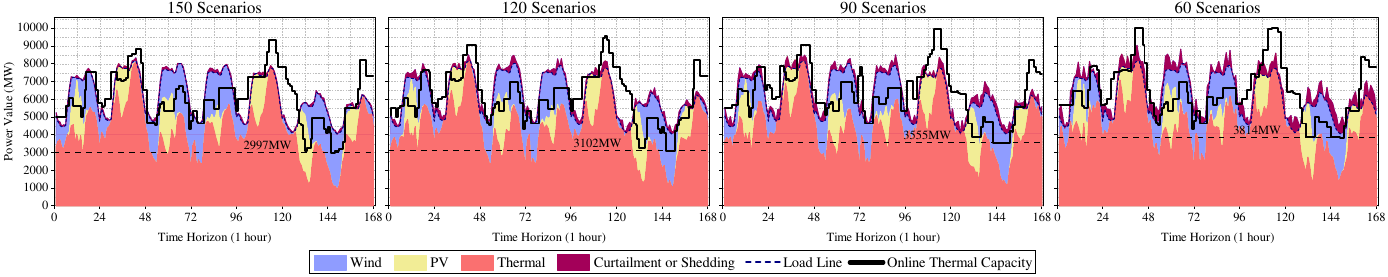}}
		\caption{Real-time power generation results of fixed day-ahead unit commitment decisions under different number of scenarios.}
	\end{figure*}
	\subsection{Case $\mathbf{C}$:  Scenario-Based Scheduling Analysis}
	To synthetically demonstrate the cost-based effectiveness of our forecast model, a simulation of scenario-based, day-ahead, and week-rolling unit commitment is implemented on the modified IEEE three-area RTS-96 system provided in \cite{system}. This test system incorporates 96 thermal units, 73 nodes, 51 loads, 107 lines, 10 wind farms, and 15 PV plants, where and the renewable scenarios are derived from the normalized observations and forecasts on the source GEFCom 2014. To simulate the actual power forecasting process, the training input for ScenGAN in this case relies on meteorological information, which is formulated as $X_{\mathbf{C}}^{\mathrm{info}}=[\mathring{\boldsymbol{c}}_1, 
	\mathring{\boldsymbol{c}}_2,
	...
	]^\intercal$, where  $\{\mathring{\boldsymbol{c}}_1, \mathring{\boldsymbol{c}}_2, ...\}$ indicates weather information time-series.
	
	After forecasting, the fragments of generated scenarios are performed as inputs to execute a two-stage stochastic programming model, obtaining the day-ahead on/off status decisions of thermal units and the expected cost. Once the day-ahead unit commitment decisions are determined and fixed, real-time economic dispatch is performed according to the observed power paths to compute the actual cost. To examine the overall picture of the scenario, the above process rolls on a daily basis and is performed with a week period. Obviously, the difference between the expected and actual costs intuitively describes the forecasting quality. Concretely, the total capacities of adopted thermal units, PV plants, and wind farms are 10215 MW, 4050 MW, and 1250 MW respectively, and we set the penalties of load shedding and renewable curtailment as 1000 \$/MWh and 40 \$/MWh. To focus on renewable scenario assessment, the uncertainty of the load is ignored. Fig. 8 illustrates the day-ahead generation results for real-time scheduling under different numbers of scenarios. The overall decision exhibits an acceptable utilization of RES generation, and with the increase in the number of scenarios, the consideration of nonschedulable resources improves, further optimizing the economics of day-ahead decision-making. Moreover, several GAN-based scenario generation models are also used in the decision-making process for comparison, as shown in Fig. 9. In the diagram, ScenGAN’s scenario exhibits optimal performance on both load shedding and renewable curtailment, and its expected cost is also closest to the actual cost. This is mainly because compared with the benchmarks, ScenGAN pays more attention to the accuracy of the local features, which confirms a reliable scenario set.
	\begin{figure*}[t]
		\centerline{\includegraphics[width=18cm]{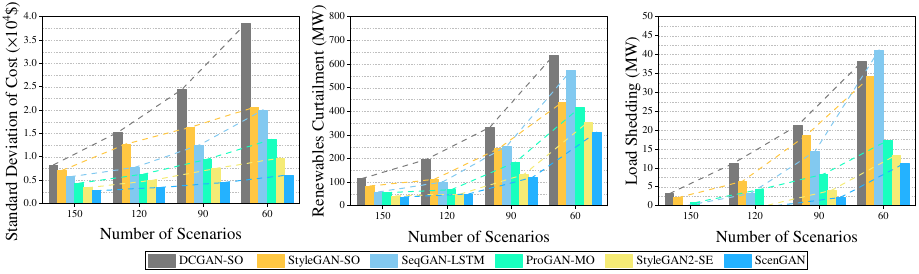}}
		\caption{Comparison of scheduling performance using different scenario models in terms of cost deviation, load shedding, and renewable curtailment.}
	\end{figure*}
	
	To compare the scheduling performance under out-of-sample conditions, a set of 500 MC realizations, 60\% of which are outside the forecasting distribution, are independently and identically sampled from the historical paths for scheduling. Table 6 exhibits the comparison results in the scheduling stage. The decisions made in ScenGAN’s scenario appear to be more robust in the face of contingencies, yielding the fewest violations and least cost variation. The main reason is the consideration of epistemic uncertainty, which enhances the generalization performance and diversified representations for the proposed model.
	\begin{table*}[h]
		\caption{Out-of-sample Comparative Testing on Day-ahead Scheduling}
		\renewcommand\arraystretch{1}
		\begin{center}
			
			\setlength{\tabcolsep}{10mm}{
				\begin{tabular}{ccccc}
					\toprule[0.5mm]
					\multirow{2}{*}{Method} 
					
					&\multicolumn{2}{c}{Cost ($\times10^6 \$$)} &  Shedding & Renewables\\
					\cmidrule(lr){2-3}
					&Average & Worst & (MW) & Curtailment (MW)\\
					\midrule
					DCGAN-SO& 2.15 & 6.42 & 7.83 & 518.64 \\
					StyleGAN-SO& 1.75 & 3.12 & 1.85 & 301.11 \\
					SeqGAN-LSTM&1.46& 4.33 & 2.78 & 374.24 \\
					ProGAN-MO& 1.41 & 2.06 & 0.81 & 97.55 \\
					StyleGAN2-SE& 1.37 & 2.28 & 0.48 & 58.52 \\
					ScenGAN & \cellcolor{Aquamarine1}\textbf{1.35} & \cellcolor{Aquamarine1}\textbf{1.89} & \cellcolor{Aquamarine1}\textbf{0.41} & \cellcolor{Aquamarine1}\textbf{34.65} \\
					\bottomrule[0.5mm]	
					
				\end{tabular}
				\label{tab1}
		}\end{center}
	\end{table*}
	\section{Conclusion}
	Focusing on the problem of large-scale uncertainties in RES generation, an uncertainty-aware scenario forecasting framework using a Bayesian deep generative model is proposed in this study. To capture epistemic and aleatoric uncertainties simultaneously, MC dropout and AdaIN are employed to increase the ability to cover the forecasting distribution. Based on the GAN architecture, an attention-intensive network generating high-fidelity scenarios based on multisource available information is designed to infer diverse time-series patterns. An analysis of three publicly available cases demonstrates that the proposed model surpasses the capacity limits of deterministic deep models and achieves outstanding performance in reliability and sharpness for renewable scenario forecasting. Follow-up work will continue to study the potential of DGMs and Bayesian theory in power systems and apply them to other challenging contexts, such as integrated forecasts considering PV-wind-load correlations and DGM-based decision-making. In addition, lightweight high-speed generation is also one of the critical technical issues.
	
	\section*{CRediT authorship contribution statement}
	\textbf{Yifei Wu:} Methodology, Writing - original draft, Software, Data curation. 
	\textbf{Bo Wang:} Software, Resources, Writing - review \& editing, Validation.
	\textbf{Jingshi Cui:} Investigation, Software. 
	\textbf{Pei-chun Lin:} Review \& editing, Validation.
	\textbf{Junzo Watada:} Review \& editing.
	
	\section*{Acknowledgement}
	This work was supported by the National Natural Science
	Foundation of China (Grant No. 61603176, 71732003), the Natural Science Foundation of Jiangsu Province (Grant No. BK20160632), and the Fundamental Research Funds for the Central Universities (Grant No. 14380037). The authors would like to express their sincere gratitude to the reviewers, whose valuable suggestions were conducive to the quality of this paper.
	
	\bibliographystyle{elsarticle-num}
	\bibliography{manuscript}
	
\end{document}